\documentclass[journal,10pt]{IEEEtran}

\usepackage{afterpage}
\usepackage{algorithmicx}
\usepackage[linesnumbered,ruled,longend]{algorithm2e}
\SetKwInOut{Input}{Input}
\SetKwInOut{Output}{Output}

\SetKwProg{Fn}{Function}{\string:}{end}

\usepackage{lipsum}
\usepackage{amsfonts} 
\usepackage{amsmath}
\usepackage{amssymb}
\usepackage{array} 

\usepackage{boxedminipage}
\usepackage{caption} 
\usepackage{cite}
\usepackage{color} 
\usepackage{enumerate}
\usepackage{float}
\usepackage{framed}

\usepackage{graphics}
\usepackage{graphicx}
\usepackage{indentfirst}
\usepackage{latexsym}
\usepackage{mathtools}
\usepackage{multirow} 
\usepackage{booktabs}
\usepackage{picinpar}
\usepackage{placeins}
\usepackage{subfigure}
\usepackage{theorem} 
\usepackage[para]{threeparttable}
\usepackage{wrapfig} 
\usepackage{textcomp}
\usepackage{verbatim}
\usepackage{multicol}
\usepackage{stfloats}

\usepackage{soul}

\usepackage{bbold}
\normalsize

\newcommand{\modify}[1]{\textcolor{black}{{#1}}}

\begin{document}

\title{Deep Reinforcement Learning Based Cross-Layer Design in Terahertz Mesh Backhaul Networks}

\author{Zhifeng~Hu, Chong~Han,~\IEEEmembership{Senior~Member,~IEEE}, and~Xudong~Wang,~\IEEEmembership{Fellow,~IEEE}
		
		\thanks{
			\par
			Zhifeng Hu, and Chong Han are with the Terahertz Wireless Communications (TWC) Laboratory, Shanghai Jiao Tong University, 200240 China (email:\{zhifeng.hu, chong.han\}@sjtu.edu.cn).
			\par Xudong Wang is with University of Michigan-Shanghai Jiao Tong University (UM-SJTU) Joint Institute, Shanghai Jiao Tong University, 200240 China (email:wxudong@sjtu.edu.cn).
		}
	}
\maketitle

\thispagestyle{empty}
\begin{abstract}
Supporting ultra-high data rates and flexible reconfigurability, Terahertz (THz) mesh networks are attractive for next-generation wireless backhaul systems {that empower the integrated access and backhaul (IAB).} 
{In THz mesh backhaul networks, the efficient cross-layer routing and long-term resource allocation is yet an open problem due to dynamic traffic demands as well as possible link failures caused by the high directivity and high non-line-of-sight (NLoS) path loss of THz spectrum.
In addition, unpredictable data traffic and the mixed integer programming property with the NP-hard nature further challenge the effective routing and long-term resource allocation design.}
In this paper, a deep reinforcement learning (DRL) based cross-layer design in THz mesh backhaul networks (DEFLECT) is proposed, by considering dynamic traffic demands and possible sudden link failures.
In DEFLECT, a heuristic routing metric is first devised to facilitate resource efficiency (RE) enhancement regarding energy and sub-array usages.
Furthermore, a DRL based resource allocation algorithm is developed to realize long-term RE maximization and fast recovery from broken links.
Specifically in the DRL method, the exploited multi-task structure cooperatively benefits joint power and sub-array allocation.
Additionally, the leveraged hierarchical architecture realizes tailored resource allocation for each base station and learned knowledge transfer for fast recovery.
Simulation results show that DEFLECT routing consumes less resource, compared to the minimal hop-count metric.
Moreover, unlike conventional DRL methods causing packet loss and second-level latency, DEFLECT DRL realizes the long-term RE maximization with no packet loss and millisecond-level latency, and recovers resource-efficient backhaul from broken links within 1s.
\end{abstract}


\section{Introduction}
\label{intro}

With the explosive growth of mobile devices accompanied by unprecedented high demands of data rates, a dramatic traffic volume is foreseen that challenges current backhaul systems, i.e., links among base stations (BSs).
Despite the ability to provide high capacity, traditional fiber-based wired backhaul is prohibitively costly and geographically inflexible~\cite{siddique2017downlink,kazemi2018wireless,al2021performance,banagar20223d}. 
Instead, wireless backhaul is a feasible alternative to guarantee cost-efficient communications everywhere, following the trend of integrated access and backhaul (IAB).
More specifically, only a few BSs (i.e., IAB donors) connect to the backbone network in a fiber-based wired manner, while the rest BSs (i.e., IAB nodes) rely on wireless communications only~\cite{huang2021bayesian,yu2023coordinated,huang2022effective}.
Unlike the conventional wired backhaul networks, similar to the access, the backhaul in IAB networks deploys the air interface as well, which is a crucial feature enabling self-adjusting topology management and plug-and-play installation~\cite{cudak2021integrated}.
To unleash the flexibility and reconfigurability of wireless backhaul for IAB in 6G and beyond systems, Terahertz (THz) band (0.1-10~THz) with 
ultra-broad available bandwidths is envisioned as a promising technology to support efficient yet practical deployment~\cite{boulogeorgos2018terahertz,akyildiz2022terahertz,sen2022multi}.


THz mesh wireless backhaul networks have the following three-fold advantages.
{First, over 20 GHz continuous bandwidth could be supplied to support multi-Gigabit-per-second IAB traffic demands~\cite{yu2020joint}.}
{Second, benefiting from large arrays of sub-millimeter-long antennas, THz line-of-sight (LoS)  multiple-input and multiple-output (MIMO) and hybrid beamforming are capable of enhancing spectral efficiency and multiplex gain, as well as alleviating interference among concurrent access and backhaul signals due to the ultra-high directivity~\cite{yan2022joint,busari2019generalized,polese2020integrated}.}
{Third, by allowing a BS to be connected with neighboring BSs, the mesh architecture enables flexible and reconfigurable backhaul topology, which makes the backhaul resilient to the changes in wireless link status~\cite{akyildiz2005survey,zhai2020mesh}.}

Even though THz mesh networks are potential to realize effective and efficient wireless backhaul systems, there are several challenges. 
First,  financial implications of the power consumption and 
the efficient sub-array management of THz hybrid beamforming are concerns~\cite{ge2018cost,zhai2019antenna}.
Second, the cross-layer routing and resource allocation design in a long-term period for resource-efficient THz backhaul is nonlinear, non-convex, and NP-hard, which is thus yet an open problem~\cite{yu2020joint}.
Third, high loss of non-line-of-sight (NLoS) paths and high directivity of THz communications might incur link degradation or failures, requiring  timely recovery to meet dynamic traffic demands~\cite{seppanen2016multipath}. 

Recently, there are efforts spent in maximizing resource efficiency (RE), defined as data rate per unit resource usage, in wireless backhaul networks.
A solution that achieves the energy efficiency maximization in backhaul with endurable complexity is proposed in~\cite{nguyen2017centralized}, by solving a beamforming and power allocation under an approximate convex problem. 
However, this method works only in a downlink, where the adopted time-division half-duplex mechanism can result in extra latency.
Another solution decomposes the NP-hard non-convex problem to separately optimize sub-array and power allocations in a convex manner, to realize the resource efficiency in terms of the number of sub-arrays as well as energy~\cite{zhai2019antenna}. 
Unfortunately, the assumptions that equal numbers of transmitting and receiving sub-arrays, together with highly predictable traffic demands are impractical. 

Apart from aforementioned non-learning methods, fueled by the recent advancement of deep reinforcement learning (DRL), learning-based solutions that can efficiently solve non-convex and NP-hard long-term  resource allocation problems have drawn much research attention recently~\cite{zhang2021cost,cui2021joint}.
Nevertheless, existing DRL solutions only focus on energy efficiency in a downlink, whereas the potential of DRL is not fully unleashed yet in low-latency bidirectional backhaul transmissions.
        \modify{In particular, the data rate requirement of the uplink is usually different from that of the downlink in backhaul networks.
        In addition, since multiple BSs can utilize the same BS as the relay on the path to the IAB donor, the antenna allocation to receive signals from multiple BSs is more complex in the uplink. 
        In contrast, in the downlink, to transmit downlink data to multiple BSs, the allocations of transmitting antennas and transmit power are more intricate. 
        In addition, the remaining resources (e.g., antennas) under the downlink resource allocation optimization may not be sufficient to support the optimization of uplink resource allocation, and cannot ensure the overall bidirectional resource efficiency as well.
        Therefore, by considering power and antenna allocations for both the uplink and downlink, as well as the different requirements of uplink and downlink data rates, the resource allocation of bidirectional transmission is much more sophisticated than the single-directional transmission. 
        As a result, specific designs of resource-efficient allocation strategies for the bidirectional backhaul transmission are motivated.}

In addition, existing non-learning and learning schemes lack the consideration of the limited storage of buffers compared to large traffic volumes, dynamic traffic demands, as well as possible link failures in THz backhaul.
These factors nullify existing non-learning convexification tricks and learning architectures, limiting the pragmatic viability of resource allocation solutions in previous works for THz backhaul networks.
Moreover, besides resource allocation considered in existing solutions, to further improve the ability of RE maximization, cross-layer design including routing should be taken into account as well.

In this paper, aiming at solving the aforementioned challenges, 
we propose a DRL based cross-layer design in a  frequency-division THz backhaul network (DEFLECT) where partial sub-bands serve the uplink, while the remainder serve the downlink.
Concretely, a heuristic solution is designed for mesh routing, 
by accounting for RE in terms of power and sub-array usage.
One step further, we develop a novel DRL based resource allocation algorithm to reach long-term power and sub-array RE maximization by jointly allocating power and sub-arrays, as well as realizing fast recovery from the broken links.
The main contributions of this paper are summarized as follows.
\begin{itemize}
    \item In a THz mesh backhaul network, pertaining to the unique attributes of THz communications, we formulate a cross-layer routing and resource allocation problem that aims at RE maximization in terms of power and sub-array usage. 
    For ease of the low computational burden for routing and handling dynamic traffic demand on a small time scale, the problem is decomposed into the routing problem and resource allocation problem.
    Physical constraints are considered, including limited power and sub-array resources, low latency, as well as packet loss mitigation.
    \item We devise a DEFLECT routing algorithm with the target of RE maximization. 
    Unlike commonly-used routing schemes that achieve the minimal hop-count~\cite{nandiraju2007wireless}, 
    an effective heuristic metric related to RE is proposed to empower the legacy Dijkstra algorithm to facilitate and expedite the RE maximization in THz backhaul networks.
    The low complexity of the proposed DEFLECT routing algorithm can realize fast recovery when link failures occur.
    \item We transform the joint resource allocation problem into an equivalent DRL problem, to achieve long-term RE maximization while satisfying physical constraints.
    In the proposed DRL framework, the multi-task architecture facilitates cooperative training for power and sub-array allocations. 
    Moreover, the hierarchical structure enables the customized resource allocation for different BSs with various numbers of links, and supports  learned information transfer for ease of fast recovery from broken links.
    {To avoid the catastrophe of severe latency and packet loss during DRL training, we further devise novel safe initialization and action exploration mechanisms in the training process.}
    \item We evaluate the performance of DEFLECT heuristic routing solution and DRL based resource allocation algorithm.
    Under different requirements of signal-to-noise-ratios (SINRs), DEFLECT routing outperforms the minimal hop-count metric in terms of  lower resource usage.
    In addition, DEFLECT DRL achieves the long-term RE maximization with no packet loss and millisecond-level latency, while conventional methods fail to avoid packet loss or second-level latency. 
    With sudden broken links, DEFLECT DRL rapidly recovers resource-efficient backhaul transmissions within 1s.
\end{itemize}

The remainder of this paper is organized as follows.
We present a THz mesh backhaul network model and formulate the cross-layer routing and joint resource allocation problem in Sec.~\ref{system}.
In Sec.~\ref{routing}, we elaborate heuristic DEFLECT routing algorithm.
DEFLECT DRL resource allocation method is detailed in Sec.~\ref{DRL discription}.
Extensive experimental results are presented in Sec.~\ref{result}.
Lastly, we conclude this paper in Sec.~\ref{conclusion}.



\section{System Model and Problem Formulation}
\label{system}

{In this section, we consider a THz mesh wireless backhaul network consisting of multiple BSs that communicate in the THz band. 
On the basis of this backhaul network model, we formulate a cross-layer routing and joint power and beamforming resource allocation problem for long-term RE maximization.}

\subsection{THz Mesh Backhaul Network Model}
\label{backhaul model}

\begin{figure}[t]
\centering
        \includegraphics[width=\linewidth]{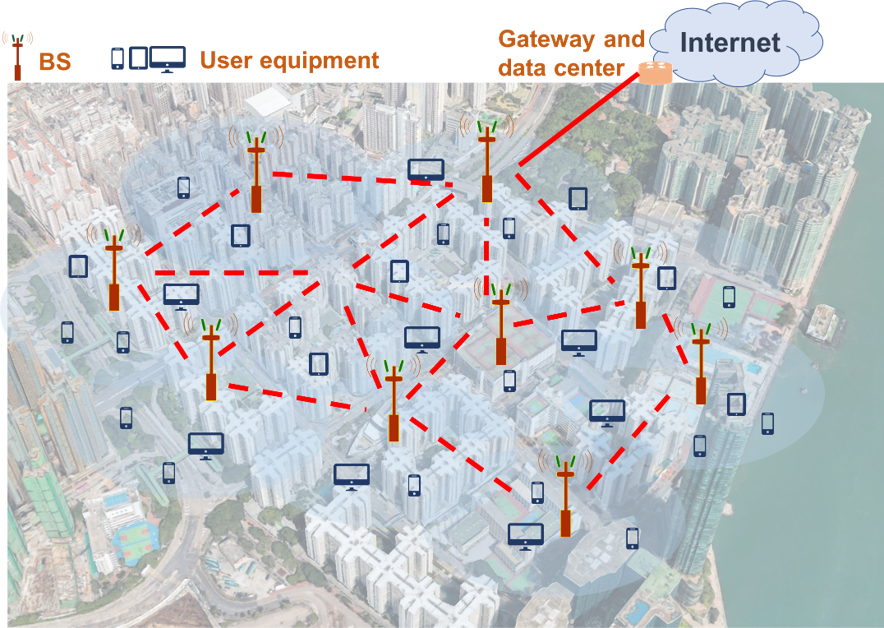} 
        \caption{{THz mesh IAB network.}}
        \label{fig:backhaul_network}
\end{figure}

{As depicted in Fig.~\ref{fig:backhaul_network}, a THz multi-hop mesh IAB network topology is adopted in this work. In particular, the backhaul links and access links are separated in the frequency domain by using different frequency bands. 
This paper is dedicated to realizing the RE maximization in the mesh backhaul model, which is described as follows.
Specifically, $N$ BS nodes comprise the backhaul network of the IAB system, in which each BS serves a cell with multiple user equipments (UEs).} 
To avoid inter-cell interference, the minimal distance between every pair of BS nodes is set as $d_\text{min}$.
For any node $i\in \{1,2,\dots,N\}$, the nodes in the proximity (i.e., within the distance of $d_\text{max}$, where $d_\text{max}>d_\text{min}$) form the neighboring set $\mathcal{N}_i$.
As a result, each BS node can communicate with other nodes in its neighboring set, i.e., within the distance in the range $[d_\text{min},d_\text{max}]$.

In addition, by considering possible LoS blockage or sudden equipment failures, some links might become inaccessible.
In this case, the failure status of this link can be quickly detected by continuously sending bidirectional probing messages~\cite{vestin2017low}, and the neighboring set thereby updates immediately by deleting the neighbor with which the node cannot communicate.

{In the mesh backhaul network, the BS node labeled as index 1 can access the gateway via fiber, which is the IAB donor, bridging the backbone network and the backhaul network. 
In contrast, other BS nodes that work as IAB nodes are assigned indices from 2 to $N$, which connect to the backbone network via wireless multi-hop links through the IAB donor.
In light of this, the backhaul topology of the IAB network is further formulated as a directed acyclic graph (DAG) to construct the link between any node from 2 to $N$ and node 1.}
The BS node~1 is therefore the root node of the DAG. 
Consequently, this DAG can be represented by an $N\times N$ adjacency matrix $A$, which is expressed as 
\begin{equation}
\label{eq:parent node}
    A_{i,j}=
    \begin{cases}
    1, &\text{if node } j \text{ is the parent node of node } i,\\
    0, &\text{otherwise},
    \end{cases}
\end{equation}
where $i,j\in\{1,2,\dots,N\}$. 
According to the definition of DAG in~\eqref{eq:parent node}, when $A_{i,j}=1, i,j\in\{1,2,\dots,N\}$, BS node $i$ transmits the uplink data to its parent node $j$, while receiving the downlink data from node $j$.

In the frequency-division THz backhaul network, the uplink and downlink traffic demands for each BS follow the fractional Brownian motion processes, which could accurately simulate the aggregated data traffic as commonly adopted in~\cite{huang2021bayesian,fidler2014guide}.
In addition, at each BS node, a first-in-first-out (FIFO) output buffer is considered.
If the buffer is full and the channel capacity cannot support transmitting more packets, any new packet arrival is dropped, which hence leads to a packet loss. 

\subsection{THz Transmission Model}
\label{transmission model}
For ease of high time efficiency and low latency, multiple sub-bands in the THz band are used by each link in a frequency-division fashion.
In particular, $K$ sub-bands are considered, where half serves for downlink transmission, and the other half takes charge of the uplink transmission simultaneously.
For each link between a pair of nodes, hybrid beamforming is adopted to magnify the signal strength as well as alleviate the interference from other links~\cite{cao2007multihop}.
Specifically for THz backhaul systems, widely-spaced multi-subarray (WSMS) architecture with enlarged inter-subarray spacing is deployed to enhance spatial multiplex gains as well as provide accurate beam alignment~\cite{yan2022joint}. 
At every BS node, the total number of sub-arrays is $S_{\max}$, and each sub-array contains $M_x\times M_y$ planar antennas.
By considering $d_0$ as the spacing of neighboring antennas in the WSMS system, array steering vector towards direction $\theta$ is given by
\begin{equation}
\begin{aligned}
    \label{eq:steeringVector}
    \mathbf{a}(\theta)=&\dfrac{1}{\sqrt{M_x M_y}}\left[1,\dots,e^{j\frac{2\pi d_0}{\lambda}(m_x+m_y)\sin(\theta)},\dots,\right.\\
    &\quad \left.e^{j\frac{2\pi d_0}{\lambda}(M_x-1+M_y-1)\sin(\theta)}\right],
\end{aligned}
\end{equation}
where $\lambda$ describes the signal wavelength, $0\leq m_x \leq M_x-1$, and $0\leq m_y \leq M_y-1$.

The MIMO channel response for the transmission from node $i$ to $j$ is
\begin{equation}
\begin{aligned}
    \label{eq:MIMOresponse}
    \mathbf{H}(i,j)=&\sqrt{(S_t(i,j)M_x M_y)(S_r(j,i)M_x M_y)}\\
    &\cdot G_tG_r\mathbf{a_r}(\theta_{r })\mathbf{a_t^*}(\theta_{t})|\alpha(i,j)|^2,
\end{aligned}
\end{equation}
where $S_t(i,j)$ and $S_r(j,i)$ denote the numbers of sub-arrays for node $i$ to transmit and node $j$ to receive. Additionally, $G_t$ and $G_r$ represent the gains of the transmitter and receiver antenna.
Based on~\eqref{eq:steeringVector}, $\mathbf{a_t}$ and $\mathbf{a_r}$ compute the steering vectors of nodes $i$ and $j$.
Moreover, $(\cdot)^*$ refers to the conjugate transpose operator, $\theta_{t }$ and $\theta_{r }$ symbolize angles of departure and arrival (AoD/AoA). 
{In addition, to capture the unique attributes in the THz channel including molecular absorption, the path gain $\alpha(i,j)$ is computed as
\begin{equation}
\label{eq:pathloss}
  |\alpha(i,j)|^2=
  \begin{aligned}
  &\left(\dfrac{c}{4\pi f d(i,j)} \right)^2e^{-g_{abs}(f)d(i,j)},
  \end{aligned}
\end{equation}
where $c$ stands for the speed of light, $d(i,j)$ means the distance between node $i$ and $j$, $f$ symbolizes the carrier frequency, $g_{abs}$ represents the medium absorption coefficient of THz signals~\cite{lin2015adaptive}.}

In the end-to-end WSMS model, the equivalent single-input and single-output (SISO) channel response of \eqref{eq:MIMOresponse} for the transmission from node $i$ to $j$  can be computed as
\begin{equation}
    h(i,j)=
    \mathbf{C_D^*}(j,i)\mathbf{C_A^*}(j,i)\mathbf{H}(i,j)\mathbf{W_{A}}(i,j)\mathbf{W_{D}}(i,j),
    \label{eq:SISOresponse}
\end{equation}
where $\mathbf{W_{A}}(i,j)$ and $\mathbf{W_{D}}(i,j)$ denote the analog and digital precoding matrices, $\mathbf{C_A^*}(j,i)$ and $\mathbf{C_D^*}(j,i)$ refer to the analog and digital \modify{combining} matrices. 

As a result, the expression on SINR for this link with the $k^\text{th}$ sub-band is
\begin{equation}
\gamma(i,j,k)=\frac{P(i,j,k)|{h}(i,j)|^2}
{|\mathbf{a^*}(\theta_r)|^2\left(I_{s}(i,j,k)+\sigma^2\right)},
\label{eq:gamma}
\end{equation}
where $P(i,j,k)$ is the transmit power assigned to the $k^\text{th}$ sub-band from node $i$, $I_{s}(i,j,k)$ indicates the interference after the self-interference cancellation and beamforming, which can be modeled as a Gaussian distribution~\cite{lei2020deep,zhang2019joint}, and $\sigma^2$ represents the noise.

The channel capacity of the $k^\text{th}$ sub-band is thereby expressed as
\begin{equation}
	\label{eq:rate}
	R(i,j,k)=\psi(i,j,k)B\log_2(1+\gamma(i,j,k)),
\end{equation}
where $\psi(i,j,k)$ is 1 or 0, indicating if the $k^\text{th}$ sub-band is utilized for the transmission from node $i$ to $j$, $B$ denotes the bandwidth for each sub-band.
{Under the WSMS physical layer design~\cite{yan2022joint}, the optimal precoding and combining for $\mathbf{W_{A}}(i,j)$, $\mathbf{W_{D}}(i,j)$, $\mathbf{C_A^*}(j,i)$, and $\mathbf{C_D^*}(j,i)$ are achievable, the channel capacity in~\eqref{eq:rate} can be reformulated as 
\begin{equation}
	\label{eq:rate multiplexing gain}
	\begin{aligned}
		R(i,j,k)=&\sum\limits_{\upsilon=1}^{\Upsilon}\left[\log_2\left(1+\frac{P_\upsilon(i,j,k)}{I_{s}(i,j,k)+\sigma^2}\varkappa_\upsilon^2 [\mathbf{H}(i,j)]\right)\right]\\
		&\cdot\psi(i,j,k)B,
	\end{aligned}
\end{equation}
where $\Upsilon=\min{\left[S_t(i,j),S_r(j,i)\right]}$ represents the multiplexing gain brought by the WSMS technique, $\varkappa_\upsilon [\mathbf{H}(i,j)]$ refers to the $\varkappa^\text{th}$ largest singular value of $\mathbf{H}(i,j)$, corresponding transmit power $P_\upsilon(i,j,k)$ that satisfies $\sum_{\upsilon=1}^{\Upsilon}P_\upsilon(i,j,k)=P(i,j,k)$ is determined by WSMS physical layer design.}

\subsection{Cross-Layer Routing and Resource Allocation Problem Formulation}
\label{problem}


In the THz mesh backhaul network, our goal is to find the optimal policy $\pi_p$ of routing (i.e., adjacency matrix $A$) and joint resource allocation (i.e., the power and sub-array allocation denoted by $\mathbf{P}$ and $\mathbf{S}$, respectively) to realize the long-term RE maximization.
Since the traffic demands are dynamic and unpredictable, while the mean values of traffic needs are given, RE is hence defined as the additive reverse of power and sub-array resource occupation.
In particular, the power occupation for node $i\in\{1,2,\dots,N\}$ is defined as \begin{equation}
    \label{eq:power occupation}
    U_P(i)=\sum\limits_{j\in\{1,2,\dots,N\}\backslash\{i\}}\sum\limits_{k\in\{1,2,\dots,K\}} \dfrac{\psi(i,j,k)P(i,j,k)}{P_\text{max}},
\end{equation}
where $P_\text{max}$ is the maximal power a BS node can provide.
The ratio of sub-array occupation among the total number of sub-arrays of $S_\text{max}$ is expressed as
        \begin{equation}
            \label{eq:array occupation}
            U_S(i)=\sum\limits_{j\in\{1,2,\dots,N\}\backslash\{i\}} \dfrac{S_t(i,j)+S_r(i,j)}{S_\text{max}},
        \end{equation}
\modify{where $S_r(i,j)$ represents the number of receiving sub-arrays for the transmission from node $j$ to $i$.}
{Under the certain values of $P_\text{max}$ and $S_\text{max}$, lower occupation ratios suggest lower resource usage (i.e., higher RE) while achieving the same rate.}
Taking into account the trade-off between RE for power and sub-array resources, the overall resource occupation is the mean of the power and sub-array occupation~\cite{tang2014resource}, given by
\begin{equation}
    \label{eq:total occupation}
    U(i)=\dfrac{U_P(i)+U_S(i)}{2}.
\end{equation}
To formulate the RE maximization problem, the objective is to minimize the expected resource occupation averaged over all BSs of the THz mesh backhaul network for a long-term period, i.e., spanning over multiple time slots $\tau$ starting from any time instant $t$, as
\begin{subequations}
        \label{eq:total objective}
        \vspace{-0.1cm}
        \begin{align}
        \label{object}
        &\mathop{\arg\min}\limits_{\pi_p\left({A_\tau, \mathbf{S_\tau},\mathbf{P_\tau}}\right)}
        \sum\limits_{\tau=t}^\infty\sum\limits_{i=1}^N\kappa^{\tau-t} \mathbb{E}_{\pi_p}\left[\dfrac{U_\tau(i)}{N}\right],\\
        \label{backbone access constraint}
        \textrm{s.t.}\quad & 
        \sum\limits_{j\in\{1,2,\dots,N\}}A_{i,j,\tau}=\left\{
        \begin{aligned}
            &0, && i=1\\
            &1, && \forall i\in\{2,\dots,N\}
         \end{aligned}\right.,\\
        \label{acyclic constraint}
         &\modify{tr\left( e^{A_\tau\circ A_\tau}\right)-N=0,} \qquad\;\! \modify{\forall \tau\geq t,}\\
        \label{power constraint}
        & U_{P,\tau}(i)\leq 1, \qquad\qquad\qquad\!\!\: {\forall i\in\{1,2,\dots,N\},}\\
        \label{sub-array constraint}
        & U_{S,\tau}(i)\leq 1, \qquad\qquad\qquad\!\!\: {\forall i\in\{1,2,\dots,N\},}\\
        \label{uplink latency constraint}
        & T^u_\tau \leq T^u_\text{max}, \qquad\qquad\qquad\,\, {\forall \tau\geq t,}\\
        \label{downlink latency constraint}
        & T^d_\tau \leq T^d_\text{max}, \qquad\qquad\qquad\,\, {\forall \tau\geq t,}\\
        \label{packet loss constraint} 
        &l_\tau \leq l_\text{max}, \qquad\qquad\qquad\quad {\forall \tau\geq t.}
        \end{align}
\end{subequations}
In \eqref{object}, $\kappa$ represents the extent to which future rewards are attenuated. Specifically, a larger $\kappa$ highlights the  
weighted contribution of future rewards, while a smaller $\kappa$ intensifies the proportion of the current reward. 
$\mathbb{E}[\cdot]$ denotes the expected value.
Since every BS node needs access to the backbone network, {DAG structure of IAB networks requires IAB nodes (i.e., nodes except BS 1) to own one parent node, while IAB donor (i.e., BS 1) directly links to the backbone network without a parent node, formulated as \eqref{backbone access constraint}.}
\modify{Moreover, in \eqref{acyclic constraint} (where $tr(\cdot)$ refers to the trace of a matrix, and $\circ$ represents the Hadamard product), the acyclicity is guaranteed, ensuring that the topology is a DAG~\cite{zheng2018dags}.}
In \eqref{power constraint} and \eqref{sub-array constraint}, the allocated power and the number of sub-arrays cannot exceed $P_{\max}$ and $S_{\max}$, which represent the maximal power and the maximal number of sub-arrays equipped at the BS.

{The latency of an uplink packet measures the time from the origin BS to arrive at BS node 1. 
Reversely, the latency of a downlink packet is the time transmitted from BS node 1 to the destination BS node.
The uplink latency $T^u_\tau$ and downlink latency $T^d_\tau$ averaged on the arrived packets cannot exceed the threshold $T^u_\text{max}$ and $T^d_\text{max}$, as shown in \eqref{uplink latency constraint} and \eqref{downlink latency constraint}, respectively. 
Moreover, the total number of lost packets $l_\tau$ cannot exceed the constraint $l_\text{max}$, as illustrated in~\eqref{packet loss constraint}.

\begin{figure*}[ht]

\newcounter{TempEqCnt} 
\setcounter{TempEqCnt}{\value{equation}} 
\setcounter{equation}{13-1} 
\normalsize
\modify{
\begin{equation}
	\label{eq:loss ij}
	l_\tau(i,j)=
        \left\{
        \begin{aligned}
            &\Gamma_\tau(i,j)-\Omega(1-O_\tau(i,j))-\left\lfloor\dfrac{R_\tau(i,j)\Delta \tau}{\omega}\right\rfloor, &&\quad\left\lfloor\frac{R_\tau(i,j)\Delta \tau}{\omega}\right\rfloor-\Gamma_\tau(i,j)<\Omega(1-O_\tau(i,j))\\
            &0, &&\quad\text{otherwise}
        \end{aligned}
        \right. .
\end{equation}\hrulefill}
\end{figure*}
\setcounter{equation}{\value{TempEqCnt}}

\modify{The packet loss and latency for the LoS transmission from node $i$ to node $j$ rely on the data rate, the number of incoming packets, and the buffer occupation state.
More specifically, in time slot $\tau$ with duration $\Delta \tau$, the data rate $R_\tau(i,j)$ is the summation of the rates over all sub-bands, which is calculated as $R_\tau(i,j)=\sum_{k=1}^{K} R_\tau(i,j,k)$. 
The number of the transmitted packets can be given by $\left\lfloor\frac{R_\tau(i,j)\Delta \tau}{\omega}\right\rfloor$, where $\omega$ is the packet size.
Furthermore, for the buffer with a storage size of $\Omega$ packets, the buffer can at most receive $\Omega(1-O_\tau(i,j))+\left\lfloor\frac{R_\tau(i,j)\Delta \tau}{\omega}\right\rfloor$ packets without packet loss, where $O_\tau(i,j)$ measures the ratio of buffer occupation.
Therefore, with $\Gamma_\tau(i,j)$ packets arriving at node $i$, whose next hop is node $j$, the packet loss of the transmission from node $i$ and $j$ is given by~\eqref{eq:loss ij}.}
The total number of lost packets in time slot $\tau$ is therefore expressed as 
\begin{equation}
\setcounter{equation}{\value{equation}+1}
	\label{eq:all loss}
	l_\tau=\sum\limits_{i=1}^{N} \sum\limits_{j\in\{j'|A_{i,j',\tau}+A_{j',i,\tau}=1\}} l_\tau(i,j).
\end{equation}
}

{In addition, in time slot $\tau$, when an uplink or downlink packet with index $\varpi$ is received by the BS node $i$ at time instant $\tau+\Delta\tau'$ (where $\Delta\tau'<\Delta\tau$) with $\Gamma'_{\tau,\varpi}(i,j)$ packets stored in the buffer before this packet, if $\left\lfloor\frac{R_\tau(i,j)(\Delta \tau-\Delta\tau')}{\omega}\right\rfloor\geq \Gamma'_{\tau,\varpi}(i,j)+1$, the latency of packet $\varpi$ from node $i$ to node $j$ is 
\begin{equation}
	\label{eq:latency fast}
	T_\varpi(i,j)=\dfrac{(\Gamma'_{\tau,\varpi}(i,j)+1)\omega}{R_\tau(i,j)}+\dfrac{d(i,j)}{c}.
\end{equation} 
In contrast, if $\left\lfloor\frac{R_\tau(i,j)(\Delta \tau-\Delta\tau')}{\omega}\right\rfloor< \Gamma'_{\tau,\varpi}(i,j)+1$ and $\Gamma''_{\tau,\varpi}(i,j)<\Omega$ (where $\Gamma''_{\tau,\varpi}(i,j)$ is the number of incoming packets after packet $\varpi$ in time slot $\tau$), packet $\varpi$ is not lost in time slot $\tau$, and its latency from node $i$ to node $j$ is expressed as
\begin{equation}
    \label{eq:latency slow}
    \begin{aligned}
           T_\varpi(i,j)=&(\mho+1)\Delta\tau-\Delta\tau'+\dfrac{(\Gamma'_{\tau+\mho+1,\varpi}(i,j)+1)\omega}{R_{\tau+\mho+1}(i,j)}\\
           &+\dfrac{d(i,j)}{c}, 
    \end{aligned}
\end{equation}
where $\mho$ satisfies that for any time slot $\tau'\in \left\{\tau+1, \tau+2,\cdots, \tau+\mho\right\}$, 
$\left\lfloor\frac{R_{\tau'}(i,j)\Delta\tau}{\omega}\right\rfloor< \Gamma'_{\tau',\varpi}(i,j) +1$, while $\left\lfloor\frac{R_{\tau+\mho+1}(i,j)\Delta\tau}{\omega}\right\rfloor\geq\Gamma'_{\tau+\mho+1,\varpi}(i,j)+1$.
Moreover, to ensure packet $\varpi$ is not lost in any time slot $\tau'$, $\Gamma''_{\tau,\varpi}(i,j)+\sum_{\tau'=\tau+1}^{\tau+\mho} \Gamma_{\tau'}(i,j)<\Omega$.
The total latency of uplink or downlink packet $\varpi$ is the summation of $T_\varpi(i,j)$ along its entire propagation path in the backhaul network.
On the contrary, the latency of packet $\varpi$ cannot be acquired if it is lost.
}

{Unlike providing throughput requirements in~\cite{zhai2019antenna}, we constrain the latency and packet loss in~\eqref{uplink latency constraint},~\eqref{downlink latency constraint}, and~\eqref{packet loss constraint}, owing to the consideration of limited buffer storage and unpredictable dynamic traffic demands.
In particular, the latency and packet loss for a packet at any time instant depend on the routing, resource allocation, buffer occupation, as well as data traffic of all BS nodes during the entire transmission time of every packet. 
Hence, the latency and packet loss constraints incur prohibitively high complexity for traditional non-learning algorithms, including exhaustive search.
In addition, to ensure that non-learning algorithms satisfy constraints~\eqref{uplink latency constraint},~\eqref{downlink latency constraint}, and~\eqref{packet loss constraint}, exact values of incoming data traffic (i.e., $\Gamma_\tau(i,j)$, $\Gamma'_\tau(i,j)$, and $\Gamma''_{\tau,\varpi}(i,j)$) are required to be obtained before routing and resource allocation solutions for every time slot.
However, this requirement is unrealistic for the unknown and unpredictable data traffic in practical backhaul networks, resulting in the infeasibility of non-learning algorithms.
Therefore, aiming at solving the above issues, DRL is adopted in our backhaul design, which can leverage powerful learning ability to support an effective solution.
}

\section{Heuristic THz Backhaul Routing Solution}
\label{routing}
Since traffic demands for each BS node follow a random process, as mentioned in Sec.~\ref{backhaul model}, the buffer occupation ratios as well as the minimal resources required to maintain efficient transmissions are dynamic. 
In contrast to the dynamic traffic demands of BSs, in light of stable backhaul transmissions and small computational and operational burdens, the mesh network topology is static and only changes when a link failure occurs with low probability.
Hence, the problem in~\eqref{eq:total objective} is decomposed into the routing problem and resource allocation problem based on the following justifications:
\begin{itemize}
    \item First, the dynamic nature of traffic demands motivates the solution for the long-term optimization of joint power and sub-array resource allocation. 
    Considering that this problem is NP-hard in mesh backhaul networks, due~to the mixed integer programming nature~\cite{zhai2019antenna}, DRL can be leveraged to allocate resources adaptively on a small time scale 
    for the long-term objective.
    Furthermore, the static topology enables the computation of routing once on a large time scale, 
    for ease of the computational burden reduction. 
    \item 
    Second, to benefit the convergence of learning, the DRL architecture requires fixed numbers of input and output for power and sub-array resource allocations.
    Indeed, these numbers depend on the number of links decided by routing.
    If routing is simultaneously updated with the resource allocation, the DRL architecture changes per training step and lacks experience in training unfortunately.
    Consequently, the routing solution needs to be firstly determined so that it can provide adequate experience for the DRL method to acquire an efficient resource allocation policy.
\end{itemize}
After the problem decomposition, the routing problem becomes
\begin{equation}
\label{eq:routing objective}
\begin{aligned}
&\mathop{\arg\min}\limits_{A_\tau}
\sum\limits_{\tau=t}^\infty\sum\limits_{i=1}^N\kappa^{\tau-t} \mathbb{E}_{\pi_p}\left[\dfrac{U_\tau(i)}{N}\right],\\
        \mathrm{s.t.}\quad&\eqref{backbone access constraint} \modify{,\eqref{acyclic constraint}.} 
\end{aligned} 
\end{equation}

Since the routing algorithm cannot predict dynamic traffic demands and corresponding resource allocation policy, routing algorithms independent of traffic demand prediction and resource allocation are motivated.
For this purpose, based on the classical Dijkstra algorithm, a heuristic scheme is leveraged to derive the solutions tailored to our routing problem in~\eqref{eq:routing objective}.
{Specifically, under the DAG structure presented in Sec.~\ref{backhaul model}, the uplink and downlink share the same routing path due to the channel reciprocity.
The aforementioned frequency-division mechanism in Sec.~\ref{transmission model} can alleviate the interference between the uplink and downlink signals.}
In light of RE maximization problem, the routing paths should incur to utilize less power and sub-array resources, while satisfying both the uplink and downlink traffic demands.
According to the relationship among power, sub-array, and distance described in~\eqref{eq:MIMOresponse},~\eqref{eq:pathloss},~\eqref{eq:SISOresponse},~\eqref{eq:gamma},~and~\eqref{eq:rate}, for a certain traffic demand of a link, the consumption of the power and sub-array resources is roughly inversely proportional to the quadratic form of the distance of the link.
Furthermore, the number of hops of each path should be small as well, since the traffic demands of each node can result in an extra transmission burden on every relay.
Consequently, unlike the hop-count metric with cost 1 for the LoS link directly connecting node $i$ to node $j$, the proposed cost metric further takes into account the factor of the distance in addition to the number of hops, which is devised as 
\begin{equation}
    \label{eq:cost}
    \varsigma_{i,j}=\iota\left[\dfrac{d(i,j)}{d_\text{min}}\right]^2 +1,  \forall i,j\in \{ 1,2,\dots,N\},
\end{equation}
where $\iota$ measures the weight of the factor of distance.
The solution is therefore to find the path from each node to the gateway with the smallest cost in~\eqref{eq:cost}.
{Although the problem in \eqref{eq:routing objective} is an integer programming problem, the low-complexity Dijkstra algorithm can be directly applied as presented Algorithm~1, since the metric in \eqref{eq:cost} satisfies the prerequisite for a positive path cost.}

\begin{algorithm}[t]
\caption{DEFLECT Routing.}
\KwIn{
The neighboring set for all node: $\mathcal{N}=\{\mathcal{N}_i|i\in \{ 1,2,\dots,N\} \}$ with the cost of the route between each pair of neighboring nodes $\varsigma_{i,j}, \forall i,j\in \{ 1,2,\dots,N\} $ in~\eqref{eq:cost}
}
\KwOut{
Adjacency matrix: $A$
}

Initialize an all zeros default adjacency matrix: $A$

Initialize an vertex set $V\leftarrow \{1, 2,\dots, N\}$

\For{each $v\in V$}
{
\If{$v=1$}{Set cost to the node 1: $C_v\leftarrow 0$}
\Else{$C_v\leftarrow \infty$}
}

\While{$V$ is not empty}
{
    $v\leftarrow$ vertex in $V$ with minimal  $C_v$
    
    Remove $v$ from $V$
    
    \For{each neighbor $u\in\mathcal{N}_v$ where $u\in V$}
    {
    \If{$C_u>C_v+\varsigma_{v,u}$}
    {
    $C_u\leftarrow C_v+\varsigma_{v,u}$
    
    $A_{u,v}\leftarrow 1$
    }
    }
    
}

\end{algorithm}

\section{Deep Reinforcement Learning Based Joint Power and Beamforming Allocation}
\label{DRL discription}

Based on the routing results provided in Sec.~\ref{routing}, the joint power and sub-array resource allocation problem can be rearranged as
\begin{equation}
\label{eq:resource allocation objective}
\begin{aligned}
&\mathop{\arg\min}\limits_{{\pi_p}\left( \mathbf{S_\tau},\mathbf{P_\tau}|A_\tau\right)}
\sum\limits_{\tau=t}^\infty\sum\limits_{i=1}^N\kappa^{\tau-t} \mathbb{E}_{\pi_p}\left[\dfrac{U_\tau(i)}{N}\right],\\
        \mathrm{s.t.}\quad&\eqref{power constraint},\eqref{sub-array constraint},\eqref{uplink latency constraint},\eqref{downlink latency constraint},\eqref{packet loss constraint}.
\end{aligned}    
\end{equation}
This problem is still NP-hard with the mixed integer programming property. 
{In addition, as discussed in Sec.~\ref{problem}, non-learning algorithms are incapable of solving this problem under the consideration of latency, packet loss, as well as unpredictable data traffic.}
Hence, we design a DRL architecture to solve the joint power and sub-array allocation in THz backhaul networks with dynamic traffic demands. 

\subsection{Deep Reinforcement Learning Framework}
\label{DRL framework}

In the context of THz mesh backhaul networks, resource allocation policy is generated by BS nodes, which perform as the DRL \textit{agent}.
In particular, 
each BS node perceives estimated SINRs and buffer occupation status of all links connected with neighboring nodes.

As an overview, the DRL algorithm operates in the THz mesh backhaul network as follows. 
Based on the current allocation policy $\pi_p$ of the actor inside each BS in each training step, each BS node processes SINRs and buffer occupation ratios observed from current state $s_\tau$ of the THz mesh backhaul environment.
Then, an action $\Xi_\tau$ is provided to allocate the transmitting sub-arrays, receiving sub-arrays, and transmit power for both the uplink and downlink.
As a result, the instant reward $r_\tau$ related to RE in~\eqref{eq:resource allocation objective} can be obtained by BS nodes.
By learning the experience of allocation actions and RE rewards, BS nodes intelligently adjust the resource allocation action for reward maximization, which is tantamount to overall resource occupation minimization.
Specifically, key components of the leveraged DRL framework for THz mesh backhaul networks, i.e., state, action, and reward, are detailed as follows.

\subsubsection{State}
\label{state}
According to~\eqref{eq:rate}, the channel capacity, which determines if the traffic demands can be satisfied, depends on SINRs.
In addition, empty buffer occupation ratios reveal that previously allocated resources are sufficient to meet traffic demands.
In contrast, when buffer occupation is not 0, more resources should be assigned to avoid the packet loss and latency caused by waiting in buffers.
To meet traffic demands, we thereby deploy the current observed SINRs for all backhaul links $\boldsymbol{\gamma}_\tau$ as well as the buffer occupation ratios for all BS nodes $\boldsymbol{O}_\tau$ as the state strategy~\cite{RLbook}, which is expressed as
\begin{equation}
    \label{eq: state}
    s_\tau=\left\{\boldsymbol{\gamma}_\tau,\boldsymbol{O}_\tau\right\}.
\end{equation}

\subsubsection{Action}
\label{action}
We devise action $\Xi_\tau=\left\{ \boldsymbol{S}_\tau, \boldsymbol{P}_\tau \right\}$ as follows.
\begin{itemize}
    \item $\boldsymbol{S}_\tau$ is a set whose elements represent the quotients of occupied sub-arrays as
    $\left\{ \boldsymbol{S}_\tau(1), \boldsymbol{S}_\tau(2), \dots,  \boldsymbol{S}_\tau(N) \right\}$.
    Specifically, 
    $\boldsymbol{S}_\tau(i)=\left\{ S_{t,\tau}(i,j), S_{r,\tau} (i,j) | j\in\{1,2,\dots,N\}, A_{i,j,\tau}=1\right\}$ denotes occupied sub-array ratios for node $i\in\{1,2,\dots,N\}$ for the transmission with all connected nodes $i$.
    To ensure the sub-array constraint in~\eqref{sub-array constraint}, 
    $\sum_{j\in\{1,2,\dots,N\}, A_{i,j,\tau}=1} \left(S_{t,\tau}(i,j)+S_{r,\tau} (i,j)\right)\leq 1, \forall i\in\{1,2,\dots,N\}$, where $ S_{t,\tau}(i,j)\geq0, S_{r,\tau}(i,j)\geq0$.
    In addition, at least one sub-array is required for the backhaul transmission.
    Hence, one sub-array is pre-allocated to each link.
    The products of the ratios and the rest numbers of sub-arrays are rounded down to yield the final sub-array allocation result.
    
    \item In the set $\boldsymbol{P}_\tau$, entry $ \boldsymbol{P}_\tau(i)=\left\{ P_{\tau}(i,j) | j\in\{1,2,\dots,N\}, A_{i,j,\tau}=1\right\}$ refers to the ratios of utilized transmit power for node $i\in\{1,2,\dots,N\}$.
    Similarly, to guarantee the power constraint in~\eqref{power constraint}, $\sum_{j\in\{1,2,\dots,N\}, A_{i,j,\tau}=1} P_{\tau}(i,j)\leq 1, \forall i\in\{1,2,\dots,N\}$, where $ P_{\tau}(i,j)\geq0$. 
\end{itemize}

\subsubsection{Reward}
\label{reward}

Aiming at the long-term RE maximization (i.e., resource occupation minimization), the instant reward $r_t$ is proportional to the additive inverse of current resource occupation (i.e., RE).
Moreover, in addition to the power and sub-arrays constraints in~\eqref{power constraint} and~\eqref{sub-array constraint} guaranteed via the action designs in~Sec.~\ref{action}, the latency and packet loss constraints in~\eqref{uplink latency constraint}, \eqref{downlink latency constraint}, and~\eqref{packet loss constraint} should be taken into consideration as well.
Hence, the penalty terms pertaining to latency as well as the number of packet loss are introduced into the reward.
As a result, the reward is expressed as
\begin{equation}
\centering
    r_\tau=-\left[
    \chi_1 \sum\limits_{i=1}^N\dfrac{U_\tau(i)}{N} + 
    \chi_2 T_\tau^u+
    \chi_3 T_\tau^d+
    \chi_4 l_\tau
    \right]
    ,
\label{eq:reward with panelty}
\end{equation}
where $\chi_1$ is a scaling factor to regulate the range of the output of the neural network for ease of the convergence.
$\chi_2,\chi_3$, and $\chi_4$ symbolize the weights of the penalties for the uplink latency, downlink latency, and packet loss, respectively.
\modify{For the purpose of practical implementation, the DRL training is strictly required to avoid large latency and packet loss in case of intolerable quality of service in the entire training process. 
However, the state-of-the-art constraint reinforcement learning solutions require a large number of episodes to reduce the actions with constraint violations~\cite{xu2021crpo,marchesini2022exploring,wu2020dynamic,murti2022constrained}.
Furthermore, experimental results demonstrate that these solutions cannot strictly ensure zero constraint violation after convergence. 
Therefore, instead of applying constrained DRL solutions, $\chi_2,\chi_3$, and $\chi_4$ can be set as large values to encourage DRL to strictly avoid the large latency or a large amount of packet loss (i.e., DRL immediately stops to explore the action with constraint violation when it encounters any constraint violation) during the entire training process on the fly.}

Consequently, in DRL context, the long-term resource-efficient resource allocation problem is equivalent to the long-term reward maximization problem in~\eqref{eq:resource allocation objective}, expressed as
\begin{equation}
    \centering
    \label{eq:DRL problem}
    \begin{aligned}
    \mathop{\arg\min}\limits_{{\pi_p}\left(\Xi_\tau|A_\tau\right)}
\sum\limits_{\tau=t}^\infty\kappa^{\tau-t} \mathbb{E}_{\pi_p}\left[
r_\tau
\right].
     \end{aligned}
\end{equation}

\subsection{Deep Reinforcement Learning Based Backhaul Resource Allocation Algorithm}
\label{All drl}

To handle the long-term backhaul resource allocation problem in~\eqref{eq:DRL problem}, we propose a DEFLECT DRL algorithm on the basis of the aforementioned framework.
By considering the action of allocation for continuous power and sub-array ratios, as mentioned in Sec.~\ref{action}, DEFLECT DRL evolves from the deep deterministic policy gradient (DDPG) algorithm that is a classical actor-critic DRL method generating continuous actions~\cite{lillicrap2015continuous}.
In the proposed DEFLECT DRL, a multi-task and hierarchical architecture is employed in each node to intelligently assign the ratios of power and sub-array usage for all links connected with neighboring BS nodes, as well as realize fast resource-efficient backhaul transmission recovery when a link failure occurs.
The detailed implementation of the DEFLECT DRL framework, actor, critic, as well as algorithm are elaborated in the following. 

\subsubsection{Overall DEFLECT DRL Framework}
\label{debar framework}

\begin{figure}[t]
\centering
        \includegraphics[width=\linewidth]{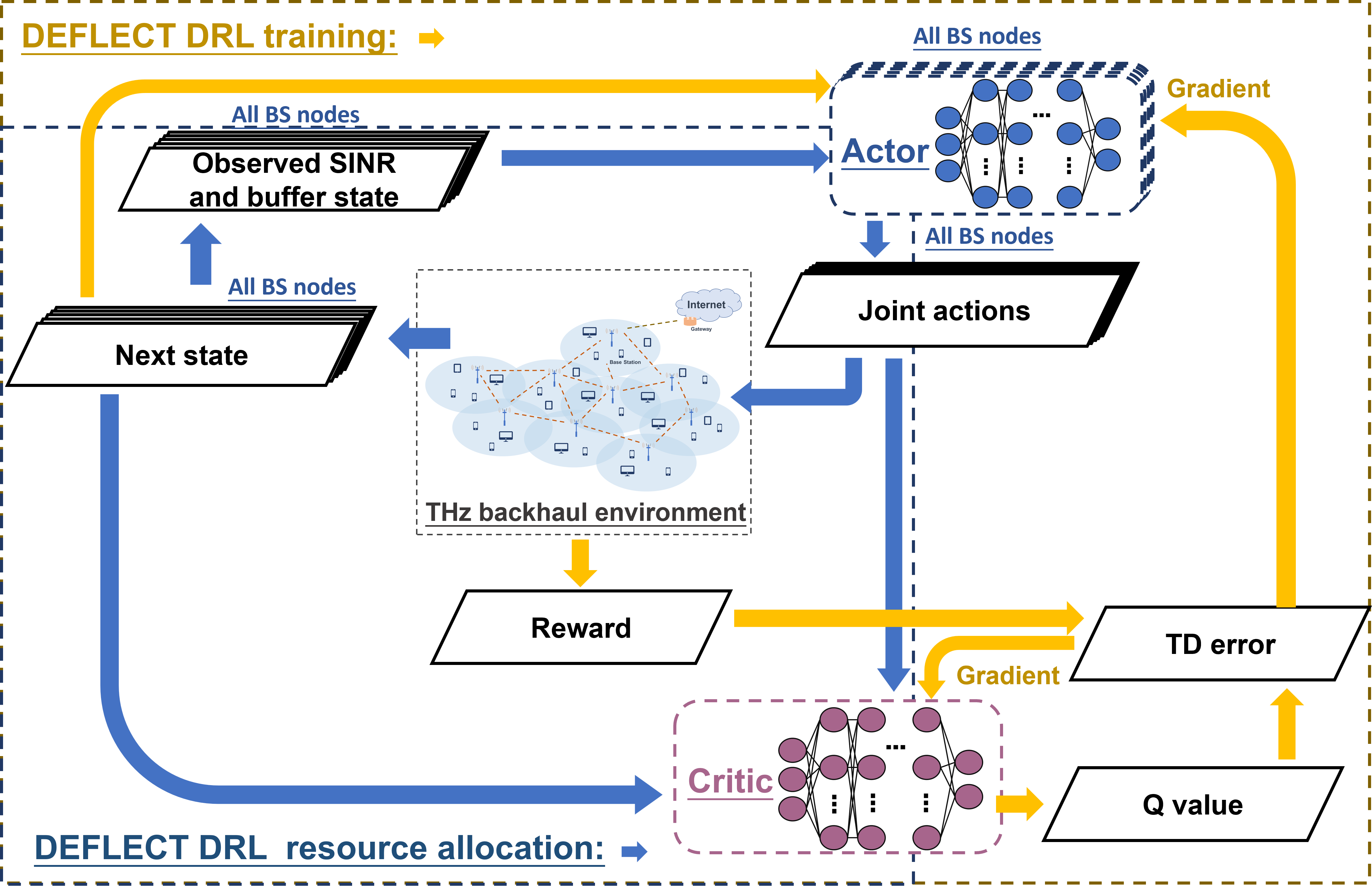} 
        \caption{{DEFLECT DRL framework.}}
        \label{fig:DeflectDRL}
\end{figure}

As depicted in Fig.~\ref{fig:DeflectDRL}, DEFLECT DRL works by cooperatively training the actor and critic.
Particularly, given the SINRs of all sub-channels and buffer occupation ratios of all links, the actor takes charge of the allocation of power and sub-array ratios for each BS node in the THz backhaul network.
In contrast, the critic outputs the Q value to evaluate the allocation generated by the actor.
Based on the resource allocation actions as well as the ceaseless uplink and downlink traffic demands, the state of SINRs and buffer occupation ratios are updated with THz backhaul traffic in Sec.~\ref{backhaul model}.

To reduce the resource waste during training, fast convergence is needed.
Furthermore, dynamic traffic demands require DEFLECT DRL to update the resource allocation policy in time. 
To this end, each BS deploys a customized actor, while the citric is employed in the data center in the backbone network. 
To cooperatively train multiple actors in all BS agents, the framework of the proposed DEFLECT DRL is based on the multi-agent DDPG~\cite{lowe2017multi} which is designed for continuous action generation in multi-agent scenarios. 
Since we only need to utilize a single Q value to evaluate the system reward in~\eqref{eq:reward with panelty}, the central critic has one output neuron for the Q value.
Consequently, the backhaul network is able to train the actors and critic simultaneously.
Instead of the sum of training time, the allocation policy given by the actor can be updated after the training time for just an actor, which expedites the training time. 
In addition, each BS node should deliver the state (i.e., SINRs and buffer occupation ratios), the actions (i.e., resource allocation for all nodes), as well as the instant reward (i.e., RE) defined in the Sec.~\ref{DRL framework} to the critic in the data center, in each training time slot. 
For a BS node with $N_c$ child nodes, at most $72(N_c+1)$ bytes should be transmitted for these data under the float32 data type, which is quite low. 
Hence, narrow-band feedback links~\cite{lu2009simple} can be applied to handle the very low feedback load.
The critic hence can be trained to give a more accurate assessment of long-term RE.
Moreover, the backpropagation information of the critic is fed back to the actor in every training step to adjust the action of allocation to achieve a higher Q value (i.e., higher long-term RE).

Aiming at feasibly implementing DRL algorithm in THz backhaul networks, the learning algorithm should provide the resource allocation result, collect the instant reward, as well as update the allocation policy on the fly.
To be practical, since only one sample experience of resource allocation actions, SINR and buffer states, as well as the reward of RE can be collected in every allocation step, 
rare experience can be collected during training, especially when fast convergence is required.
To solve this requirement, on-policy training is adopted, to make the algorithm focus on the resource allocation optimization according to the current state in the THz backhaul environment with dynamic traffic demands. 
Particularly, at every time slot, the previous state, actions, reward, and current state are fed into DEFLECT DRL for training.
Then, based on the current parameters, DEFLECT DRL generates the resource allocation actions by processing the current state, which are directly applied to the THz mesh backhaul network.
In addition, these actions, state, reward, as well as next state are utilized as the training data in the next time slot.

\subsubsection{Actor}
\label{actor}

\begin{figure*}[t]
\centering
        \includegraphics[width=\linewidth]{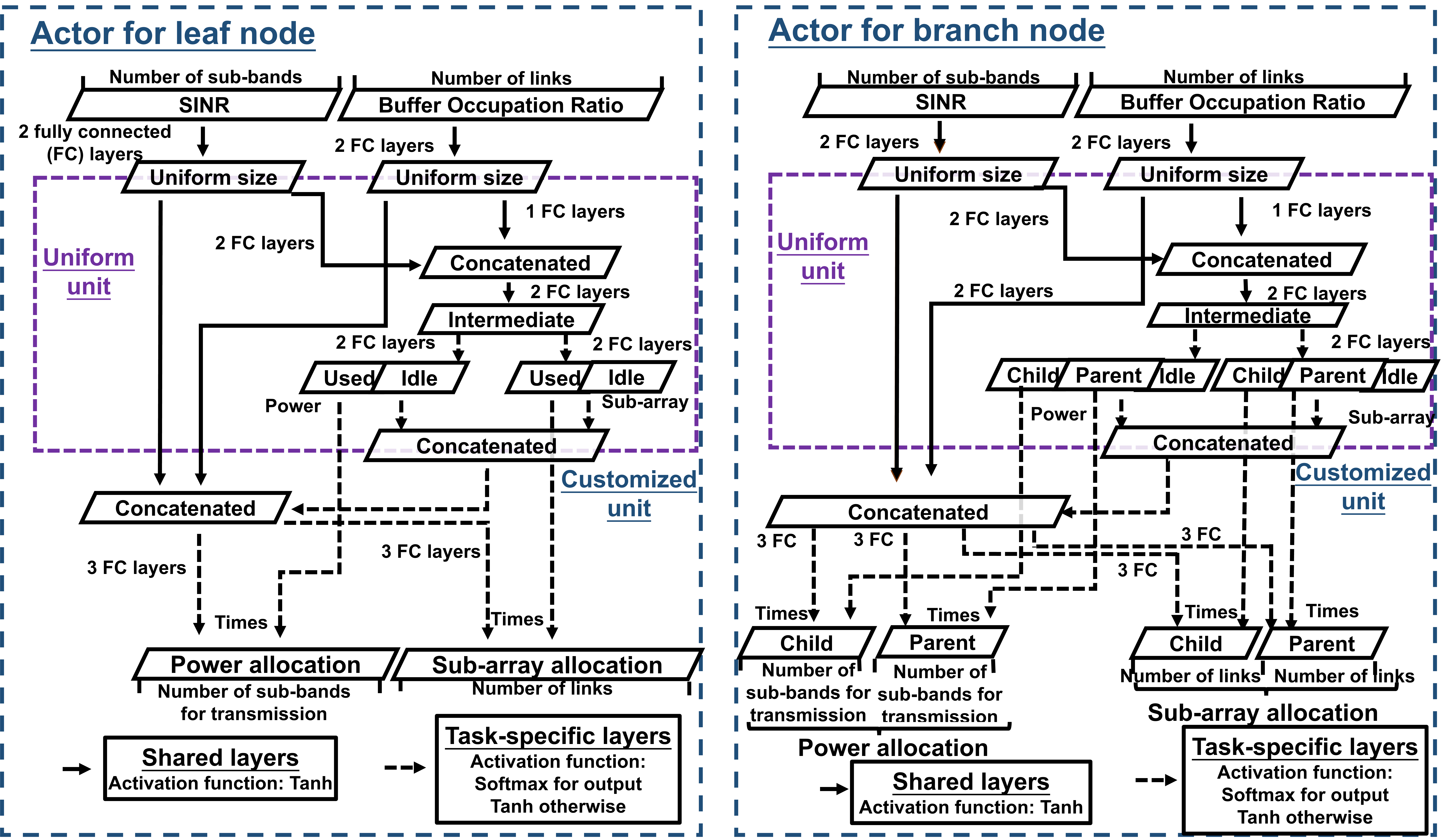} 
        \caption{{Architecture of the DEFLECT DRL actor.} }
        \label{fig:actor}
\end{figure*}

\modify{At each BS node, an actor observes SINRs and buffer occupation status of each link connected with parent and child BS nodes.
By processing the partially observed state via neural networks, this actor determines the continuous ratios of transmitting and receiving sub-arrays for the links connected with other nodes, as well as the transmit power ratio for each sub-band.}
As illustrated in Fig.~\ref{fig:actor}, the actor leverages the multi-task and hierarchical architecture to synchronously train the joint power and sub-array resource allocation and rapidly recover the backhaul transmission when any link failure happens, respectively.
In particular, shared and task-specific layers comprise the multi-task structure that divides the action of joint allocation into power allocation and sub-array allocation.
Since both the power and sub-array allocation rely on the input state of SINRs and buffer occupation ratios, the shared layer first explores the input state and extracts features.
Additionally, the shared layers can be trained by the backpropagation of both power and sub-array allocations, which prompts the actor to find vital features in observed states. 
The relationship between different allocation tasks can be jointly learned and utilized as well.
Then, the task-specific layers process these features and exploit the valuable information for each task.
The power and sub-array allocation tasks are generated by the task-specific layers simultaneously according to the attributes of power and sub-array allocation defined in Sec.~\ref{action}, aiming at the long-term RE maximization.

Furthermore, for ease of fast resource-efficient THz backhaul transmission recovery, the hierarchical structure exploited in DEFLECT DRL contains a uniform unit and a customized unit.
Specifically, on account of the fast recovery of backhaul transmissions, the uniform units for the actors in different BS nodes own  identical sizes of input and output. This brings convenience in transferring learned parameters, when a link failure occurs and the routing changes.
In addition, the uniform units determine the ratios of idle power and sub-array resources, which are directly related to RE in~\eqref{eq:reward with panelty}.
\modify{Moreover, to ensure that the power and sub-arrays constraint in~\eqref{power constraint} and~\eqref{sub-array constraint} are satisfied, the output layers of the uniform units utilize the softMax as the activation function. 
Particularly, for any input numbers $\{z_1,z_2,\dots,z_\rho\}$, the output of softMax function is given by
\begin{equation}
\label{eq:softmax}
    softMax(z_{\rho'})=\dfrac{e^{z_{\rho'}}}{\sum_{\rho''\in \{1,2,\dots,\rho\}}e^{z_{\rho''}}}, \forall \rho'\in\{1,2,\dots,\rho\}.    
\end{equation}
As a result, the summation of occupied and idle sub-arrays or power ratios is 1.
Hence, by subtracting the idle ratios, the output values of utilized sub-array or power ratios are upper bounded by 1.}
On the contrary, the numbers of input and output of the customized unit for each BS node depend on the number of links of this BS node. 
\modify{Through the customized unit, the utilized power and sub-array ratios given by the uniform unit are further distributed to each sub-channel and each link via softMax function, respectively.}
As illustrated in Fig.~\ref{fig:actor}, unlike BS nodes called leaf nodes without child nodes, other BS nodes called branch nodes also relay packets for some other nodes.
Hence, the allocation actions take into account the power and sub-arrays used for communicating with not only the parent nodes but the child nodes for branch nodes. To compare, for leaf nodes, only the resources for the links with the parent nodes are considered.

\subsubsection{Critic}
\label{critic}

\begin{figure*}[t]
\centering
        \includegraphics[width=\linewidth]{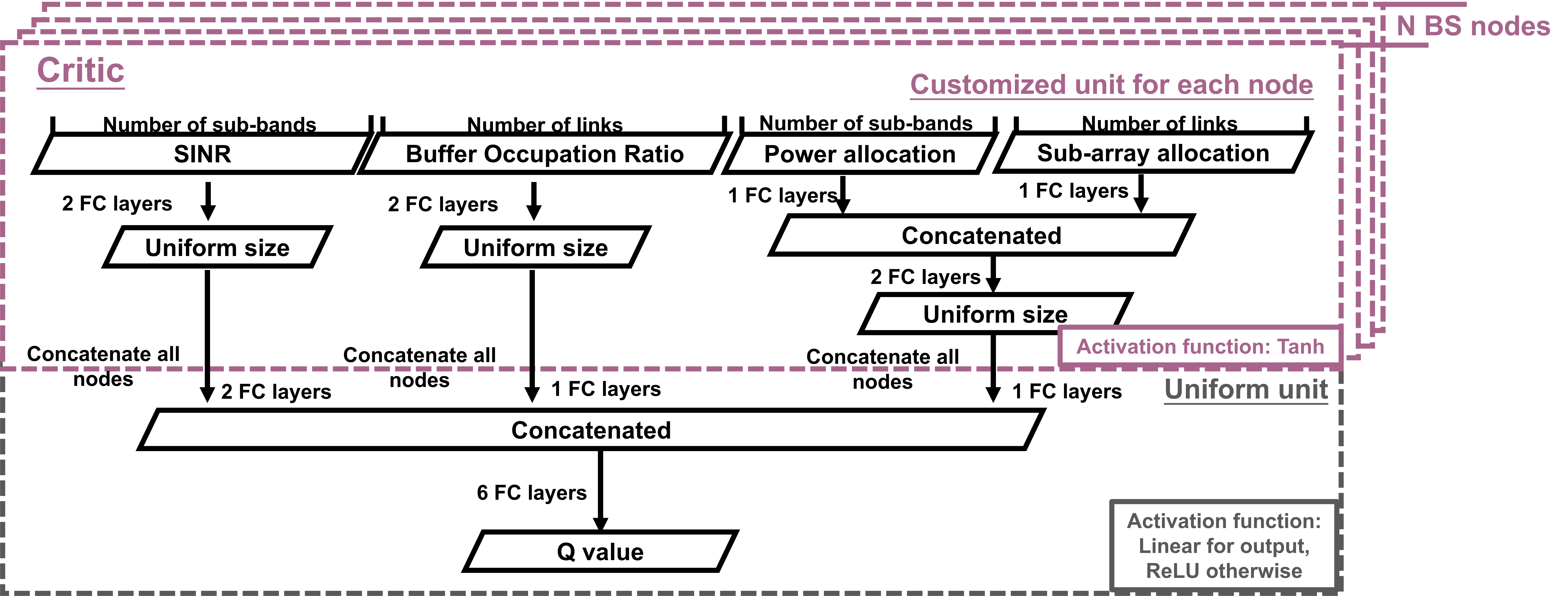} 
        \caption{{Architecture of the DEFLECT DRL critic.}}
        \label{fig:critic}
\end{figure*}

Apart from the actor in each BS node that assigns power and sub-array resources, the critic located at the data center provides Q values to jointly assess the actions of power and sub-array allocation given by the actor.
Similar to the actor, the critic also utilizes a hierarchical architecture to explore the observed SINR and buffer states as well as resource allocation actions, as depicted in Fig.~\ref{fig:critic}.
More specifically, for each node, the observed SINR and buffer states, as well as resource allocation actions, are processed by a customized unit into features with the identical sizes similar to that in the actor.
Then, these features are concatenated and fed to the uniform unit, which generates the Q value measuring the potential maximal accumulated reward (i.e., RE) as
\begin{equation}
\label{eq:Q}
Q(s_t,\Xi_t)=\max\limits_{\left\{\Xi_\tau|\tau>t\right\}}\mathbb{E}\left[\sum\limits_{\tau=t}^{\infty}\kappa^{\tau-t} r_\tau|s_t,\Xi_t\right],
\end{equation}
for any SINR and buffer state $s_t$ and allocation action $\Xi_t$.

\subsubsection{DEFLECT DRL Algorithm}
\label{debar algorithm}
The DEFLECT DRL training process is summarized in Algorithm 2, and described here.
In the context of DEFLECT DRL framework, the actions of resource allocation to realize the resource-efficient THz backhaul transmission is tantamount to reaching the maximal Q values.
In light of this, the Q values of the critic should be accurate.
Particularly, the optimal Q value, $Q^*$, satisfies the Bellman equation, which is defined as 
\begin{equation}
\label{eq:bellman function}
\centering
    Q^*(s_\tau,\Xi_\tau)=\mathbb{E}[r_t+\kappa Q^*(s_{\tau+1},\Xi'_{\tau+1})|s_\tau,\Xi_\tau],
\end{equation}
where $\Xi'_{\tau+1}$ is the action to realize the highest $Q^*(s_{\tau+1},\Xi'_{\tau+1})$.
Hence, the loss function of the critic with parameters $\theta_c$  should minimize the temporal difference (TD) error of Q values, as
        \modify{\begin{equation}
        \begin{aligned}
                \label{eq:loss critic}
            Loss(\theta_c)=\mathbb{E}\Big[\Big(r_\tau+\kappa Q(s_{\tau+1},\Xi_{\tau+1}|{\theta_c}) -Q(s_\tau,\Xi_\tau|\theta_c)\Big)^2\Big],
        \end{aligned}
        \end{equation}
        where $\Xi_{\tau+1}$ is the action of the next time slot given by the current policy of the actor, on the basis of the next state $s_{\tau+1}$.}
        According to gradient descent, $\theta_c$ is updated with the learning rate $\lambda_c$ for the ${i'}{}^\text{th}$ training step as
        \modify{\begin{equation}
        \begin{aligned}
                \label{eq:critic update}
               \theta_c^{i'+1} =&\theta_c^{i'}-\lambda_c\nabla_{\theta_c} Loss(\theta_c^{i'})\\
               =&\theta_c^{i'}-2\lambda_c\mathbb{E}\Big[\Big(r_\tau+\kappa Q\Big(s_{\tau+1},\Xi_{\tau+1}|{\theta_c}\Big)
               \\&-Q(s_\tau,\Xi_\tau|\theta_c)\Big)\nabla_{\theta_c}Q(s_\tau,\Xi_\tau|\theta_c)\Big].
        \end{aligned}
        \end{equation}
        }
Based on the accurate Q values, the objective function of the actor with parameters $\theta_a$ is to select actions achieving the Q value maximization, which is expressed as
\begin{equation}
\label{eq:loss actor}
\centering
    J(\theta_a)=Q(s_\tau,\Xi_\tau|\theta_c). 
\end{equation}
{By applying gradient ascent with learning rate $\lambda_a$, the update of the actor is therefore attainable as
\begin{equation}
\begin{aligned}
        \label{eq:actor update}
       \theta_a^{i'+1} =& \theta_a^{i'}+\lambda_a \nabla_{\theta_a}J(\theta_a) \\
       =&\theta_a^{i'}+\lambda_a\nabla_{\theta_a}Q(s_\tau,\Xi_\tau|\theta_c)\\
       =&\theta_a^{i'}+\lambda_a\nabla_{\Xi_\tau}Q(s_\tau,\Xi_\tau|\theta_c)\nabla_{\theta_a}\Xi_\tau.
\end{aligned}
\end{equation}
}

In addition, since DEFLECT DRL is trained on the fly for practical deployment, the allocation that triggers large latency or a large amount of packet loss might jeopardize the backhaul transmission.
In addition, DRL learns from collected experience.
DRL might not timely adjust the actions of allocation well before it thoroughly learns all the actions causing such latency and packet loss problems, which leads to a catastrophe in the THz backhaul network.
To preclude these issues, novel \textit{safe initialization} and \textit{safe exploration} mechanisms are devised in the proposed DEFLECT DRL as follows. 
\begin{itemize}
    \item To avoid that DEFLECT DRL learns from experience with latency and packet loss problems in the beginning, in which case DEFLECT DRL cannot know which actions are good, initial resource allocation should utilize almost all resources. 
    To this end, in the actor of DEFLECT DRL, weights for generating the ratios of idle resources are initially modified to produce values close to 0. 
    \modify{In particular, this modification is feasible due to the specific architecture of the proposed uniform unit in each actor, which uses the softMax function in~\eqref{eq:softmax} to determine the amount of utilized and idle resources.
    When we initialize the output layer of each actor neural network, the weights are set as zero, while the biases are set as negative numbers.
    Consequently,  actor networks of DEFLECT DRL are guaranteed to provide near-zero idle resources with the Softmax activation function. 
    Different from conventional neural networks without safe initialization, as well as the aforementioned state-of-the-art safe mechanisms~\cite{xu2021crpo,marchesini2022exploring,wu2020dynamic,murti2022constrained} that require a large number of episodes to avoid any constraint violation,  DEFLECT can ensure low latency and no packet loss from the beginning.} 
    \item Aiming at enabling the actor to explore actions beyond local optimum, Gaussian random noises are adopted when the allocation actions generated by the actor are deployed on the THz backhaul BS nodes.
    Specifically, the noises are added to every element of allocated power and sub-array ratios, as well as  the ratios of idle resources.
    Since the ratios of idle resources are critical for RE, to realize efficient action exploration, the variance of the Gaussian noise for the ratios of idle resources is 5 times larger.
    In addition, to assure the power and sub-array constraints in~\eqref{power constraint} and~\eqref{sub-array constraint}, the means of the sum of noise are forced into zero.
    When the random noise might cause any allocated ratios to be negative ratios, the risk that some links are incompetent in satisfying traffic demands arises.
    In that case, the action noises are withdrawn, and the action generated by the actor is directly employed.
\end{itemize}

\begin{algorithm}[t]
\caption{DEFLECT DRL Training.}
\KwIn{
Actor network parameters: $\theta_a$; Critic network parameters: $\theta_{c}$
}
\KwOut{
Well trained network parameters
}

Initialize SINR and buffer state $s_t$

Initialize $\theta_a$ to allocate all the resources

    \For{$\tau=t,t+1,\dots$}
    {
        Select action $\Xi_\tau$ with random action noise according to the state $s_t$ and the policy $\pi_p$ generated by actor 
        
        Calculate reward $r_\tau$ with $s_\tau$ and $\Xi_\tau$
        
        Update $s_{\tau+1}$ 
        
        Calculate $\Xi'_{\tau+1}={\pi_p}(s_{\tau+1}|{\theta}_a)$
        
        Calculate $
        y_\tau=r_\tau+Q(s_{\tau+1},\Xi'_{\tau+1}|{\theta}_{c})
        $
        
        $\theta_a$ $\leftarrow$ Adam gradient ascent $Q(s_\tau,\Xi_\tau|\theta_{c})$ 
        
        $\theta_{c}$ $\leftarrow$ Adam gradient descent $
        [y_\tau-Q(s_\tau,\Xi_\tau|\theta_{c})]^2$
    }

\end{algorithm}

To move one step forward, the proposed DEFLECT DRL can recover RE when broken links suddenly arise.
Owing to the design of uniform units of the actor and critic, whose input and output sizes are fixed, the learned information can be transferred into the new actor and critic that are designed for the changed numbers of links of some BS nodes.
Particularly, the uniform unit of the actor takes charge of determining the ratios of idle resources by exploring the features of observed SINR and buffer state. This directly relates to RE and hence reserves the important information for the training of RE maximization.
Moreover, the uniform unit of the critic directly outputs the Q value by processing the state and action features.
Consequently, the proposed DEFLECT DRL is able to realize fast resource-efficient THz backhaul transmission recovery via the following process, and the corresponding process is illustrated in Algorithm 3.

When a link is suddenly broken, the routing is recomputed first as Algorithm 1.
Then, according to the updated routing, the actor of each BS and critic reinitialize their customized units.
Meanwhile, the critic, as well as each actor of BS nodes that maintain the node type (i.e., branch or leaf node), reserves the parameters of the uniform unit.
By contrast, each actor of BS nodes that change types updates its uniform unit by utilizing well trained parameters average on uniform units of all the nodes with the corresponding node type.

\begin{algorithm}[t]
\caption{{DEFLECT Recovery.}}
\KwIn{
Original actor network parameters: $\theta_a$; Original critic network parameters: $\theta_{c}$; Detecting broken link in the original routing result
}
\KwOut{
Actor network parameters after the link failure
}

Recompute the routing as Algorithm 1 

\For{Node $i=1,2,\dots,N$}
{
Reinitialize the customized unit of actor

\If{Node $i$ changes types (including branch node and leaf node) from $\varrho_1$ to $\varrho_2$}
{
Transfer parameters of original uniform units average on nodes of $\varrho_2$ to the uniform unit of actor
}
\Else
{
Reserve original parameters of the uniform unit of actor
}
}

Reinitialize the customized unit of critic

Reserve original uniform unit of the critic

Conduct DRL training as Algorithm 2

\end{algorithm}

\section{Simulation Results and Analysis}
\label{result}

{In this section, we conduct experiments to assess the performance of DEFLECT regarding the factors considered in DRL rewards in~\eqref{eq:reward with panelty}, including RE, latency, and packet loss.
In comparison, the benchmark schemes are the classical minimal hop-count routing metric and conventional DRL methods (i.e., deep Q-learning (DQN) and actor-critic (AC) algorithms).}

\subsection{Simulation Parameters of THz Mesh Backhaul Networks}

\label{sec: drl results}
\begin{table}[tp]
\centering
\caption{Hyperparameters in DEFLECT DRL.}
\begin{tabular}{@{}lc@{}}
\toprule
\textbf{Hyperparameter}                        & \textbf{Value}                              \\ \midrule
Attenuating factor for future rewards $\kappa$ & 0.5                                  \\
Training slots                                  & 200                                  \\
Variance of action noises                      & $5\%\times$ maximal allocated ratios \\
Learning rate                                  &  $\lambda_a=0.03,\lambda_c=0.1$      \\
Scaling factor $\chi_1$                        & 100                                  \\
Penalty weight for uplink latency $\chi_2$     & 5000                                 \\
Penalty weight for downlink latency $\chi_3$   & 5000                                 \\
Penalty weight for packet loss $\chi_4$        & 0.1                                  \\ 
\bottomrule
\end{tabular}
\label{tb:hyper}
\end{table}

\begin{figure}[t]
\centering
        \includegraphics[width=.5\linewidth]{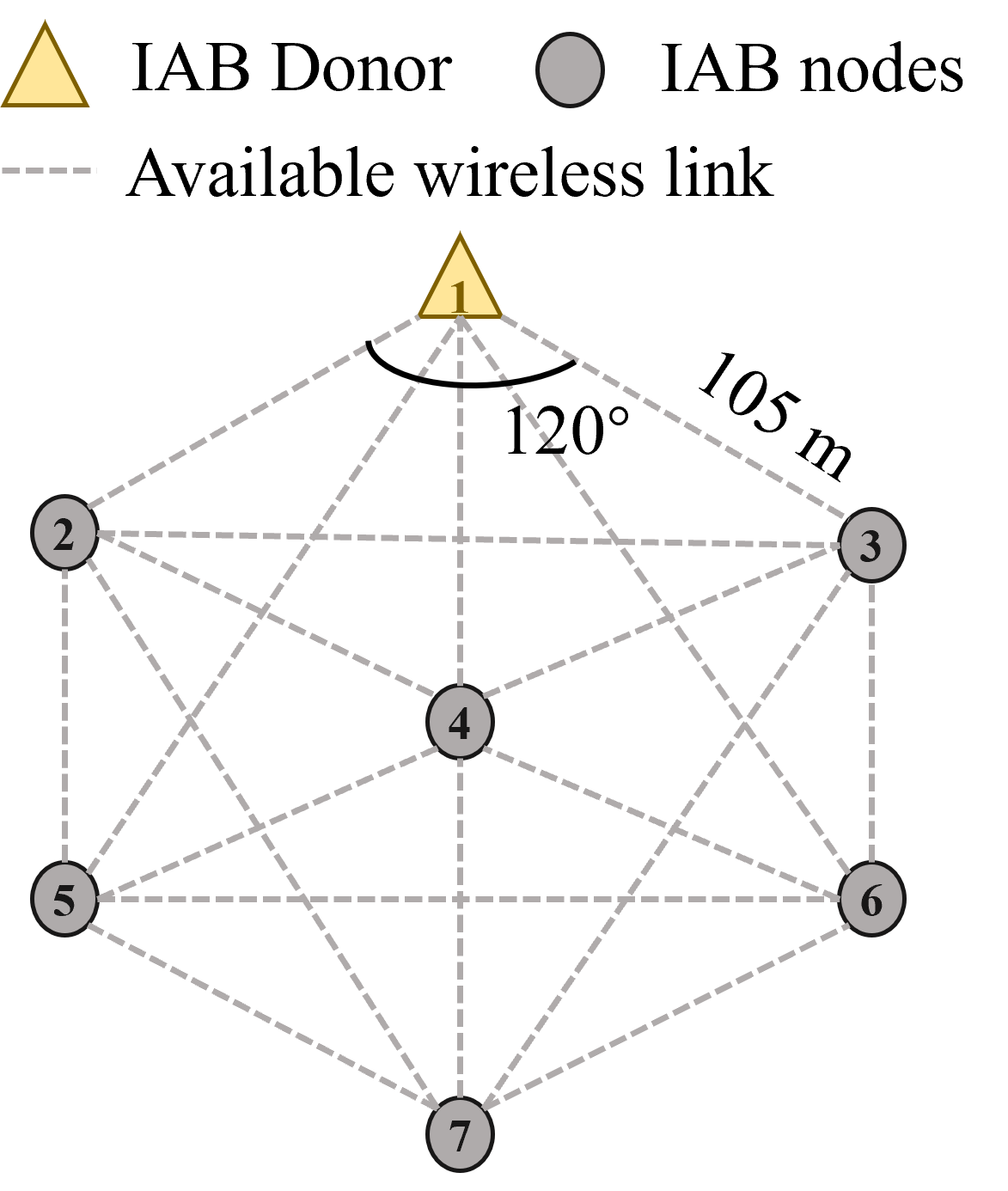} 
        \caption{{Hexagonal backhaul topology for THz mesh IAB network.}}
        \label{fig:topology}
\end{figure}

We conduct experiments on the basis of Pytorch 1.8.2 and Python 3.6.13 running on AMD Ryzen ThreadRipper 3990X CPU. 
The experiments are established on a classical regular hexagonal THz mesh backhaul network, as shown in Fig.~\ref{fig:topology}.
In particular, each BS can communicate with neighboring BSs within a distance range of $[d_\text{min},d_\text{min}]=[100~\text{m},200~\text{m}]$.
In addition, each BS deploys $p_\text{max}=30~\text{dBm}$ transmit power and $S_\text{max}=64$ sub-arrays, where each sub-array consists of $M_x\times M_y=4\times4$ planar antennas.
Specifically, the omnidirectional antennas are adopted with $G_t=G_r=0$~dB.
According to penalties of latency and packet loss constraints in DRL problem in~\eqref{eq:reward with panelty} and~\eqref{eq:DRL problem}, the latency and number of packet loss should be minimized together with power and sub-array usage.
Five nonoverlapping sub-bands with 5~GHz bandwidth within 275--300~GHz are allocated for uplink transmissions.
Similarly, five sub-bands selected from 300--325~GHz are used for downlink transmissions.

By considering the maximal running time for each training step and resource allocation decision as 0.13s in the measurement of our experiments, the time interval between two consecutive DRL actions is set as $\Delta \tau=0.15$s.
Moreover, $\Delta \tau$ is adopted as the time for a training slot, in which DEFLECT DRL is trained once. 
For each BS, the means for uplink and downlink traffic are $\mu_{up}=2\times10^4$ and $\mu_{dn}=5\times10^4$~packets in each training slot, respectively.
{In particular, each packet of backhaul transmission contains $\omega=2000$ bytes of data~\cite{xia2021multi}.}
Consequently, the mean uplink and downlink traffic demands for each BS are 5.33~Gbps and 2.13~Gbps.
Furthermore, each output buffer can store at most $\Omega=2\times10^5$ packets.

\subsection{DEFLECT Routing Performance Evaluation}
\label{sec: routing results}

\begin{figure}[t]
\centering
        \includegraphics[width=\linewidth]{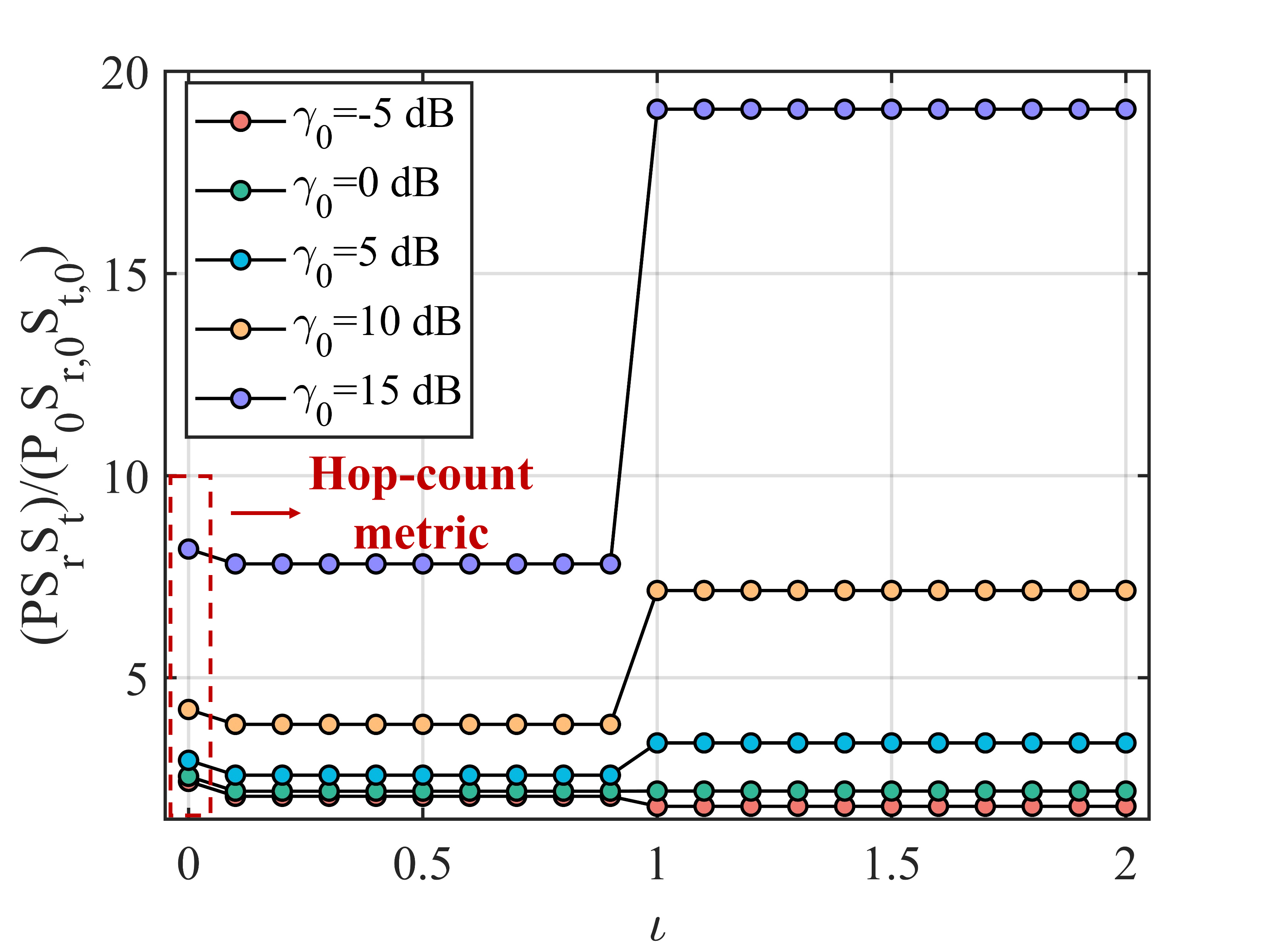} 
        \caption{Resource consumption with different weight $\iota$.}
        \label{fig:routingRE}
\end{figure}

We evaluate the performance of our heuristic routing algorithm.
Since the routing results are static unless broken links occur, we ideally assume that all BS nodes own the same static traffic demands in the analysis.
As shown in~\eqref{eq:MIMOresponse},~\eqref{eq:SISOresponse},~\eqref{eq:gamma}, and~\eqref{eq:rate}, the data rate directly relates to the product of the transmit power $P$, the number of transmitting sub-arrays $S_t$, and the number of receiving sub-arrays $S_r$.
Hence, we analyze the resource consumption in terms of $PS_tS_r$ of downlink transmission.
Noticeably, uplink transmission has the same results as the downlink transmission due to the same relationship between data rate and resource occupations.
Intuitively, the routing result with lower $PS_tS_r$ is more likely to realize lower resource consumption in~\eqref{eq:total occupation} and hence higher RE.

With bandwidth of $B_0$, the downlink traffic demand for each BS node can be expressed as $R_0=B_0\log_2\left(1+\gamma_0\right)$, where $\gamma_0$ is SINR required for leaf BS nodes without relaying packets from other BS nodes.
We demonstrate $PS_tS_r$ averaged on each BS node under different $\gamma_0$ in Fig.~\ref{fig:routingRE}, by considering that a leaf BS node consumes $P_0$ transmit power, $S_{t,0}$ transmitting sub-arrays, and $S_{r,0}$ receiving sub-arrays to support traffic demand $R_0$ with distance $d_\text{min}$.
{As depicted, different weights of distance $\iota$ in the path cost metric in~\eqref{eq:cost} result in different routing paths.}
When $\gamma_0<0$~dB, less consumption of $PS_tS_r$ is reached if $\iota>1$.
On the contrary, $\iota\in(0,1)$ results in less consumption  for $\gamma_0>0$~dB.
Interestingly, when $\gamma_0=0$~dB, any $ \iota>1$ realizes the same  $PS_tS_r$.
Moreover, when $ \iota=0$, the metric is equivalent to the hop-count metric without considering the distance, which never provides the highest RE for any value of $\gamma_0$.
In particular, taking $\gamma_0=0$~dB as an example, the optimal $\iota$ realizes 15.4\% less resource usage, compared to the hop-count metric.

{To enable timely recovery from the link failure, routing for THz mesh backhaul networks is required to run fast.
With the assistance of the Fibonacci heap, our routing algorithm inherits the low complexity of Dijkstra algorithm, which is $O\left( |E|+|N|\log |N| \right)$~\cite{fredman1987fibonacci}, where $E$ represents the number of available wireless links in the backhaul network with $N$ BSs. 
Specifically, $E$ and $N$ depend on the network topology, and $E$ is in the range of $\left[N-1,\binom{N}{2}\right].$
In particular, our routing algorithm spends $2.1\times10^{-2}$ms in the hexagonal topology, revealing its high time efficiency.}


\begin{figure}[t]
\centering
        \includegraphics[width=\linewidth]{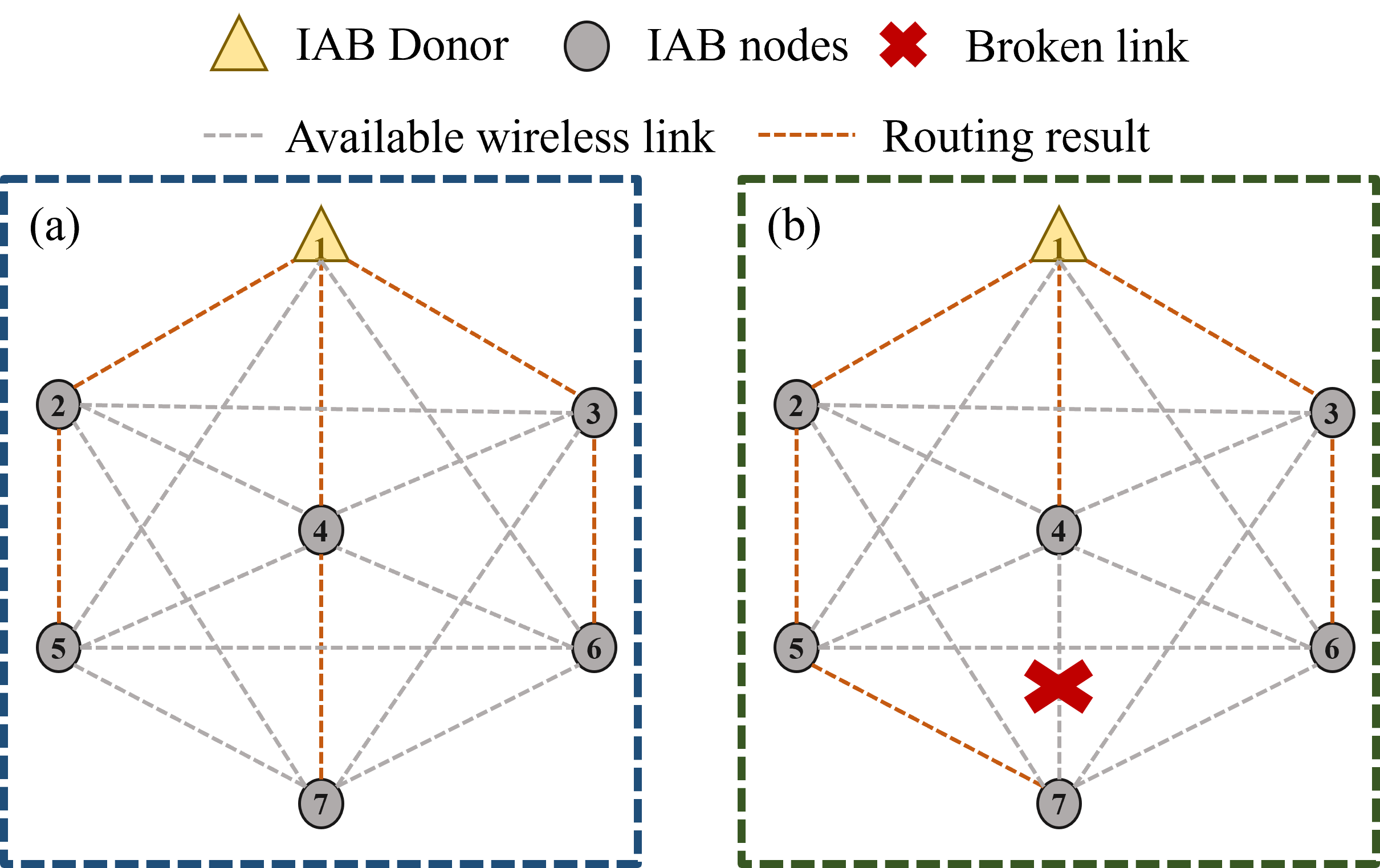} 
        \caption{{Routing results. (a) before sudden link failure, (b) after sudden link failure.}}
        \label{fig:DRLrouting}
\end{figure}
\subsection{DEFLECT DRL Performance Evaluation}


To evaluate DEFLECT DRL performance, we choose the routing metric with $\iota=1$. 
The routing results before and after the broken link are shown as Fig.~\ref{fig:DRLrouting}(a) and Fig.~\ref{fig:DRLrouting}(b), respectively. 
The hyperparameters of DELFELCT DRL are summarized in Table~\ref{tb:hyper}, which are tuned to guarantee the convergence of training.
Based on these hyperparameters, we evaluate the performance of the proposed DEFLECT DRL during training and after the link failure, 
compared with results of unsafe DEFLECT DRL, DQN, and AC. 
Specifically, the benchmark algorithms are designed as follows.
{
\begin{itemize}
    \item\modify{Unsafe DEFLECT DRLs have three types, all of which keep the same architecture as the DEFLECT DRL, while type I lacks the safe exploration mechanism described in Sec.~\ref{debar algorithm}, type II lacks the safe initialization mechanism, and type III lacks both safe exploration and safe initialization.}
    \item DQN~\cite{mnih2013playing} is a conventional DRL scheme generating discrete actions deterministically by directly estimating Q-values for all actions without critic.
    In light of this, to avoid violating power and sub-array constraints in~\eqref{power constraint} and~\eqref{sub-array constraint}, ratios of sub-arrays are picked from $\left\{0, \frac{\bar{S}}{9},\frac{2\bar{S}}{9}, \dots,\bar{S}\right\}$, where $\bar{S}$ is the ratio of sub-array usage when sub-arrays are uniformly assigned to each link.
    Similarly, each BS assigns transmit power for each sub-channel from $\left\{0, \frac{\bar{P}}{9},\frac{2\bar{P}}{9}, \dots,\bar{P}\right\}$, where $\bar{P}$ is the allocated ratio for uniform power allocation.
    In our experiments, DQN exploits the same structure as the DEFLECT actor, except for its unique output layer.
    \item 
    \modify{By exploiting the actor-critic structure, AC~\cite{sutton1999policy} is another conventional DRL method. 
        Specifically, AC leverages both the actor generating the probabilities to choose each action and the critic assessing the chosen action. 
        Since the number of output neurons is discrete, the architecture of the actor can only provide discrete actions.
        In light of this, the action space of power and sub-array allocation is the same as DQN.}
        In addition, AC adopts the same structure as the DEFLECT DRL, except for the customized output layer of the actor.
\end{itemize}
}


\begin{figure}[t]
\centering
        \includegraphics[width=\linewidth]{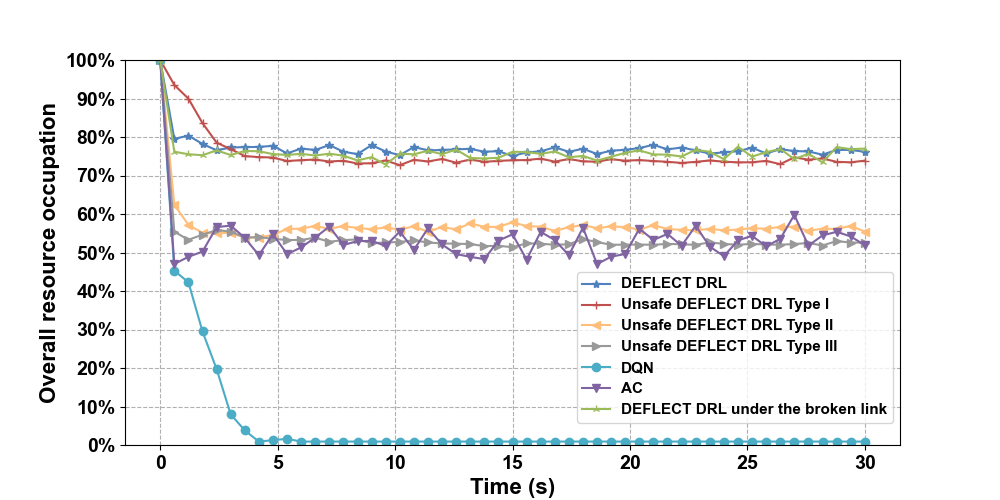} 
        \caption{Resource occupation comparison.}
        \label{fig:RE comp}
\end{figure}

\begin{figure}[t]
\centering
        \includegraphics[width=\linewidth]{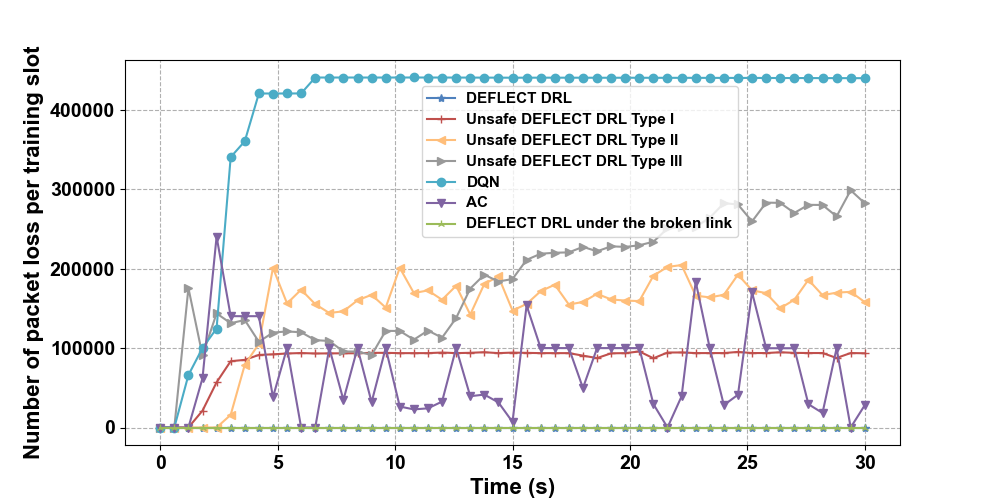} 
        \caption{Packet loss comparison.}
        \label{fig:error comp}
\end{figure}

\begin{figure}[t]
\centering
        \includegraphics[width=\linewidth]{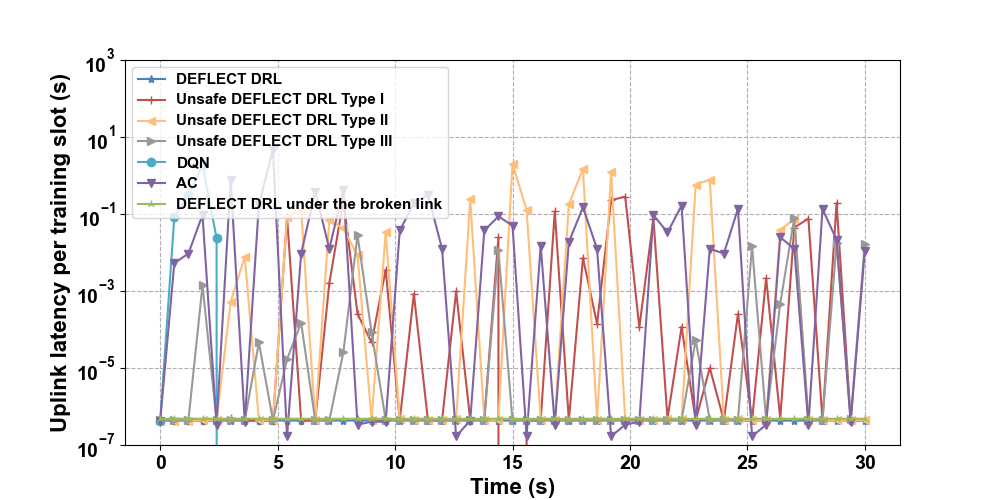} 
        \caption{Uplink latency comparison.}
        \label{fig:up delay comp}
\end{figure}

\begin{figure}[t]
\centering
        \includegraphics[width=\linewidth]{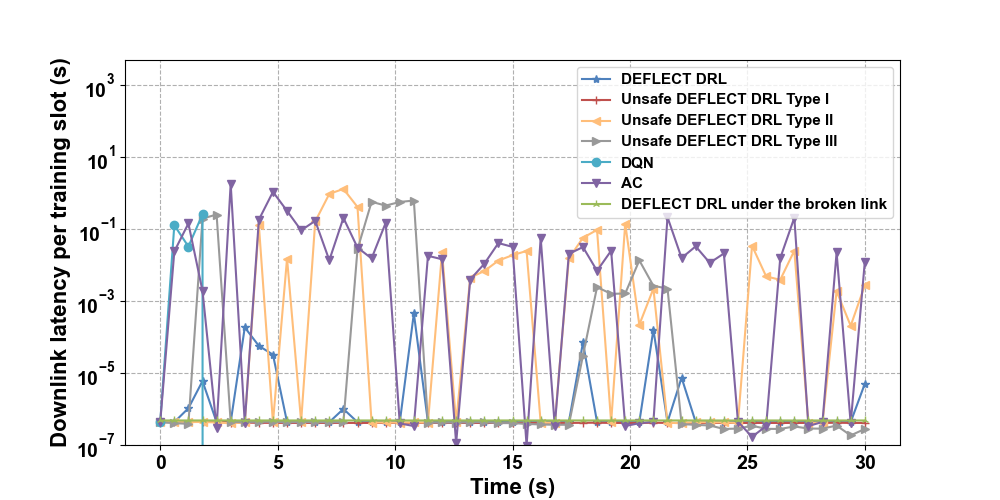} 
        \caption{Downlink latency comparison.}
        \label{fig:dn delay comp}
\end{figure}

Fig.~\ref{fig:RE comp} demonstrates the resource occupation in~(11) averaged on all BSs in training.
        As depicted, all the algorithms reduce the resource occupation and converge within 10s during training. 
        More specifically, DEFLECT DRL diminishes the overall power and sub-array resource usage from 100\% and converges at 75\% usage after 3s. 
        \modify{By contrast, unsafe DEFLECT DRLs type I, type II, and type III converge at 73\%, 55\%, and 52\% resource occupation after 5s, respectively.}
        DQN incurs to allocate no power and sub-array at convergence, and AC fluctuates in the range of $[46\%,54\%]$ resource usage.
        However, although all the schemes are trained to obtain lower resource usage, low latency and package loss before and after training convergence are significant to guarantee the quality of service.
        As illustrated in Fig.~\ref{fig:error comp}, the three benchmark algorithms reduce resource usage at the expense of severe packet loss. 
        \modify{In particular, unsafe DEFLECT DRLs type I, type II, and type III cause up to $1.0\times10^5$, $2.1\times10^5$, and $3.0\times 10^5$ lost packets per training slot, respectively.
        Therefore, safe initialization and safe exploration can both alleviate the packet loss.
        Additionally, AC incurs up to $2.4\times10^5$ lost packets per training slot, while DQN leads to more than $4.4\times10^5$ lost packets per training slot after convergence.}
Hence, they are infeasible to support practical THz mesh backhaul networks.
\modify{Interestingly, DQN prefers to send no packers with zero resource usage.
Intuitively, one major reason is that compared to AC and DEFLECT DRL, the architecture of DQN lacks a central critic, which enables cooperative training with multiple agents (i.e., BS nodes).
In particular, in the beginning (i.e., within 1s), the penalty brought by the packet loss is zero.
Hence, since the DQN has no shared critic, every agent (i.e., BS node) greedily inclines to reduce resource usage for higher rewards. 
In addition, unlike stochastic AC, DQN chooses action deterministically, which may limit its exploration of discrete actions with higher rewards as well.
As a result, each agent greedily utilizes zero resources without cooperative training.}
On the contrary, DEFLECT DRL never triggers any packet loss before and after the link failure.
This is owing to the assistance of \modify{safe exploration and safe initialization in combination}, enabling DEFLECT DRL to avoid the exploration of actions leading to packet loss.
\modify{Unfortunately, even AC outperforms the unsafe DEFLECT DRL as illustrated in Fig.~\ref{fig:RE comp} and Fig.~\ref{fig:error comp}, it cannot deploy safe initialization and safe exploration to further improve its performance due to the following two reasons.
First, AC is designed to select the actions for each BS node in a random way~\cite{sutton1999policy}, since it can only output the probabilities of action selections.
Hence, it is impossible to exploit safe initialization to force AC to utilize all resources initially.
Second, unlike DDPG, which adds noises to explore actions, AC is designed to explore the action space by selecting the random resource allocation action for each BS node.
The proposed safe exploration mechanism specifically designed for DDPG hence cannot be used for AC.
Additionally, the aforementioned discrete action spaces for AC already ensure that the power and sub-array constraints in~\eqref{power constraint} and~\eqref{sub-array constraint} are satisfied, and the values of allocated power and subarrays are non-negative.
Therefore, the exploration for AC is already safe.} 

The latency averaged on successfully arrived uplink and downlink packets are shown in Fig.~\ref{fig:up delay comp} and Fig.~\ref{fig:dn delay comp}, respectively.
As illustrated, the uplink latency for DEFLECT DRL is very low, suggesting that no uplink packet is delayed in buffers.
Since the downlink traffic is larger, DEFLECT DRL with limited buffer storage leads to millisecond-level downlink latency.
However, DEFLECT DRL timely adjusts the allocation actions, precluding packet loss and additional latency in the next training slot.
For example, DEFLECT DRL firstly encounters millisecond-level downlink latency in around 3s, in line with when it converges and stops reducing resource usage. 
In comparison, \modify{AC and unsafe DEFLECT DRL type III frequently bring second-level uplink and downlink latency}, which is much larger than DEFLECT DRL.
\modify{Unfortunately, since all types of unsafe DEFLECT DRL and DQN cause significant packet loss, the numbers of successfully arrived packets are small or even 0. 
Hence, the latency of these algorithms cannot be acquired in some training slots without successfully arrived packets, while the latency of remaining successfully arrived packets might be low in some slots.}

With occurrence of a broken link,  DEFLECT DRL rapidly converges at 75\% within 1s thanks to the heuristic structure transferring learned information before the link failure.
Furthermore, zero packet loss and low latency  before the link failure can still be maintained,  
illustrating the efficiency and reliability of proposed DEFLECT DRL algorithm.

\section{Conclusion}
\label{conclusion}
In this paper, we proposed DEFLECT for the long-term RE maximization in a THz mesh backhaul network in a cross-layer manner.
Specifically, the RE maximization problem is decomposed into a routing problem and a joint power and sub-array allocation problem, by considering practical traffic demands and possible link failures.
On one hand, a heuristic metric was proposed to enable resource-efficient DEFLECT routing.
On the other hand, we devised DEFLECT DRL to intelligently allocate power and sub-arrays in each BS for the target of long-term RE maximization.
In particular, DEFLECT DRL employs a multi-task structure to assign power and sub-arrays cooperatively.
Additionally, a hierarchical structure enables both tailored resource allocation for each BS and rapid transmission recovery from broken links.

Experimental evaluation demonstrates that DEFLECT routing can provide routing results with lower expected resource usage than the minimal hop-count metric, according to different requirements of SINRs.
In addition, DEFLECT DRL realizes long-term RE maximization with no packet loss and millisecond-level latency on the fly.
The heuristic architecture of DEFLECT DRL achieves the fast recovery of resource-efficient backhaul transmissions from broken links within 1s as well.

\modify{Furthermore, the proposed DEFLECT is promising to be implemented on backhaul networks with a large amount of BSs, due to the low-complexity Dijkstra-based routing and DRL algorithms.
Specifically, since all actors of DEFLECT DRL are trained and provide results simultaneously, the complexity of the actors can be approximately regarded as a constant if the number of BSs increases.
By contrast, the critic receives the actions and states of the actor in each BS that are linearly related to the number of parameters of the critic.
Hence, the complexity of the critic increases linearly with the increase in the number of BSs, if the hyperparameters of hidden layers are fixed. Therefore, as future work, rigorous evaluations on the practical deployment in large-scale mesh backhaul networks with possibly low-complexity algorithms will be studied.}

	\bibliographystyle{IEEEtran}
	\bibliography{bibdata}
\end{document}